\documentclass[10pt, twocolumn]{article}

\usepackage[utf8]{inputenc}
\usepackage{amsmath, amssymb, amsfonts}
\usepackage{graphicx}
\usepackage{hyperref}
\usepackage{fancyhdr}
\usepackage{geometry}
\usepackage{titlesec}
\usepackage{xcolor}
\usepackage{listings}
\usepackage{caption}
\usepackage{multicol}
\usepackage{float}
\usepackage{microtype}
\usepackage{enumitem}
\usepackage{mathtools}
\usepackage{tikz}
\usepackage{booktabs}
\usepackage{listings}
\usepackage{xcolor}
\usepackage{subcaption}

\titlespacing\section{0pt}{12pt plus 2pt minus 2pt}{8pt plus 2pt minus 2pt}

\title{\vspace{-1cm}\textbf{H-Model: Dynamic Neural Architectures for Adaptive Processing}}
\author{Dmytro Hospodarchuk \\ \texttt{hosp.dimon@gmail.com}}
\date{}

\begin{document}

\hypersetup{
    colorlinks=true,
    linkcolor=gray,
    urlcolor=blue,
    citecolor=red,
}

\onecolumn
\tableofcontents
\newpage
\twocolumn

\maketitle

\thispagestyle{fancy}

\begin{abstract}
This article explores the design and experimentation of a neural network architecture capable of dynamically adjusting its internal structure based on the input data. The proposed model introduces a routing mechanism that allows each layer to influence how its outputs are propagated through the network, enabling iterative and adaptive computation. This concept is loosely inspired by the idea of thought processes and dynamic reasoning, where information flow is conditioned not only on the data itself, but also on the internal state of the system.

It is important to note that this work does not aim to compete with state-of-the-art language models in terms of performance. Instead, it presents a conceptual prototype—an architectural framework that opens up a new direction for exploring adaptable and potentially more interpretable networks. The goal is not optimization of existing benchmarks but rather the proposal of a system that can learn not only representations, but also the structure of computation itself.

Due to practical constraints in computing resources and data, this study remains a preliminary investigation. Nevertheless, initial observations show promise, and the architecture’s full potential can only be evaluated in future experiments under more favorable computational conditions.
\end{abstract}

\section{Introduction}

The emergence of deep learning \cite{backpropagatingerrors, all_about_deeplearning_LeCun2015} has fundamentally reshaped the landscape of artificial intelligence by enabling models to automatically learn and represent complex, hierarchical patterns in data. Rather than relying on manual feature engineering, deep learning systems extract meaningful representations directly from raw inputs, which has led to breakthroughs across a wide range of domains, from image recognition and machine translation to generative modeling and strategic decision-making.

This success has largely stemmed from the increased representational power granted by deep architectures. As models have grown deeper and more expressive, they have become better at capturing the compositional structure of real-world phenomena and have gained the ability to modulate more complex patterns. However, this line of progress—deeper, wider, and more parameter-heavy models—represents only a single axis of modeling capability. While it enhances capacity, it does not necessarily introduce new dimensions of flexibility or adaptability. Thus, we believe that granting the models more and more freedom in computations is the direction that makes us closer to the desired AGI.

To move beyond this one-dimensional trajectory, recent research has begun exploring architectures that can adapt their structure or behavior dynamically. Such models aim not just to learn from data, but to reorganize their computational graph in response to the data. This shift marks a transition from static pipelines to more flexible, modular, and even self-reflective architectures—systems that decide not only what to compute, but how to compute it, depending on the input and context.

Several approaches have contributed to this paradigm:

\begin{itemize}
    \item \textbf{Dynamic Depth Models} \cite{graves2017adaptivecomputationtimerecurrent, sabour2017dynamicroutingcapsules, hsu2018unifiedmodelextractiveabstractive}: Adjusting the number of layers or computational depth based on the complexity of each input.
    \item \textbf{Mixture of Experts (MoE)} \cite{adaptive_moe, sparse_adaptive_moe}: Selecting specialized sub-networks (experts) on a per-input basis to scale capacity without scaling computation linearly.
    \item \textbf{Neural Architecture Search (NAS)} \cite{zoph2017neuralarchitecturesearchreinforcement, pham2018efficientneuralarchitecturesearch}: Automating the discovery of architecture configurations tailored to specific tasks.
    \item \textbf{Evolutionary Algorithms}: Evolving model structures over time using biologically inspired principles.
    \item \textbf{CREWAI and Dynamic Systems}: Enabling real-time communication between models.
    \item \textbf{Multimodal Transformers and Attention Mechanisms} \cite{attention, vaswani2023attentionneed, bert, vit, git, blip}: Allowing shared computation across diverse data types, such as language, vision, and audio.
    \item \textbf{Ensemble Methods}: Combining multiple models to capture a richer distribution over possible behaviors.
\end{itemize}

Each of these approaches contributes to the broader goal of increasing a model’s capacity not just in terms of depth or width, but in terms of conditional behavior, specialization, and generality. This is where our proposed architecture is positioned.

This work introduces a data-aware model that adapts its architectural structure in response to the characteristics of the input, enabling flexible and context-sensitive computation. By dynamically routing information through different parts of the network, the model becomes more capable of distributing and isolating domain-specific knowledge across its parameters, potentially leading to better generalization across diverse tasks and modalities. This introduces a meta-learning behavior, where the model effectively selects a computational pathway suited to each instance.

Conceptually, this design acts as a wrapper around traditional architectures, allowing us to create a model that takes advantage of different models on the layer level, thus we can take advantage of several large pretrained models at the same time. It introduces a learned routing mechanism alongside the usual representation learning process, allowing the model to adapt its depth, width, and information flow iteratively during inference, thus the model can decide itself on the architecture to use. While simple in implementation, this approach offers a promising route toward more flexible and potentially more efficient learning systems.

\begin{figure}[h!]
    \centering
    \begin{subfigure}[t]{0.45\textwidth}
        \includegraphics[width=\linewidth]{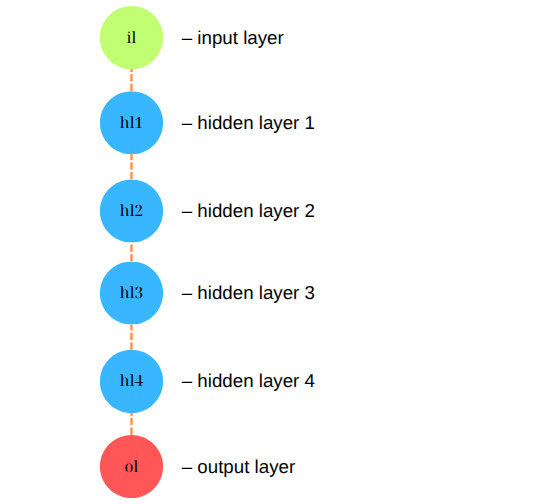}
        \caption{Traditional models}
    \end{subfigure}
    \hfill
    \begin{subfigure}[t]{0.30\textwidth}
        \includegraphics[width=\linewidth]{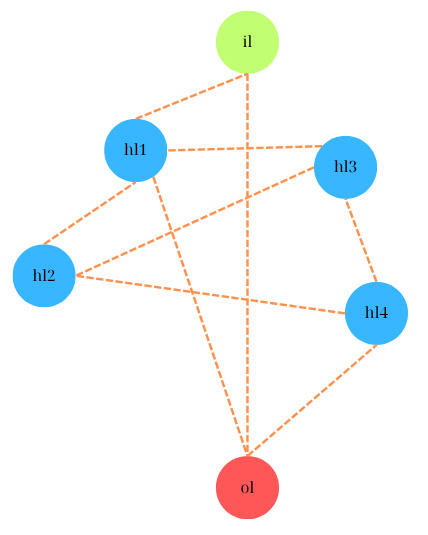}
        \caption{H-model}
    \end{subfigure}
    \caption{Comparison of conventional Forward pass and our model.}
    \label{fig:traditional_method_and_hmodel}
\end{figure}
\section{The Concept}

Before diving into the architectural details of the proposed model, it is essential to lay out the core idea behind its design—one that challenges the traditional approach to neural network computation and opens the door to more adaptive, modular, and flexible architectures.

In conventional neural networks, the flow of information is rigidly structured. Each layer performs its computation and sends its output directly to the subsequent layer in a fixed, sequential order. This forward-pass computation, where outputs from one layer feed directly into the next, has been the standard model for deep learning for years. However, this rigid structure—while highly effective in many tasks—can limit the flexibility of the model, especially when faced with complex, varied, or multimodal input data. We believe that is what limits modern models' capability to generalize more, and generalization is the core contribution to AGI. We also question modern multimodal architectures that deal with different types of data at the same time, usually by totally separate pathways for each type of input. Thus our model have the potential to better deal with modular inputs (for example, image-based qa) due to the way our meta-model shares hidden states between layers on each iteration.

The key idea behind the proposed model is to move beyond this fixed sequence of operations. Instead of treating the network as a simple, linear chain of computations, we propose a dynamic routing mechanism where each layer actively decides to which other layer (or layers) it will pass its output. This creates a flexible network where the flow of information can adapt based on the nature of the data. The network effectively "rolls up" into a dynamic, input-dependent graph, where layers can establish new, context-sensitive relationships with one another.

The metaphor we use to describe this approach is inspired by the connections between neurons in the human brain \cite{wang2014functional, human_brains_func_spec, Battaglia_2012}. In the brain, the connections between neurons aren't fixed or rigid; instead, they adapt based on the inputs and the processing needed. Similarly, in the proposed model, the layers aren't bound to pass information in a simple, pre-defined order. Each layer has the capability to dynamically choose its communication partners, effectively altering the network's structure in response to the input data.

\subsection{Dynamic Information Flow}
The model starts with a predefined input layer, just as traditional networks do. However, from that point onward, each layer in the network has the autonomy to determine how to route its output. Instead of passing its output to just one subsequent layer, a layer can send its information to multiple layers or even re-route its computation depending on the task at hand. This routing decision is learned during the training phase, allowing the model to discover the most efficient and specialized paths for different types of inputs. In this way, the model's computation is no longer a fixed chain; it becomes a dynamic process that adapts in real time.

This concept is a departure from the traditional feedforward architecture, where each layer has a fixed, predefined successor. In our model, the computation is fluid, with information flow being conditional and context-sensitive. Layers can decide, for instance, that for one input, information should flow through certain layers that specialize in processing particular types of features, while for another input, a different set of layers may be activated, potentially involving layers with different types of specialization or even combining information from multiple layers.

\subsection{Potential for Modular and Specialized Knowledge}
One of the most promising aspects of this architecture is its potential for modularity and specialization. In traditional networks, all layers are essentially homogeneous in their computational roles, and while they may learn to extract different types of features from data, they are not capable of specializing in a way that is highly flexible or context-driven. By enabling layers to dynamically route information, we introduce the possibility for a more modular design, where certain layers or sets of layers can specialize in handling specific domains or data types. For example, one set of layers might focus on processing language data, while another set might focus on visual information, with both being activated as needed, depending on the input.

Furthermore, by allowing the model to dynamically form connections, there is a potential for the model to reuse its parameters more efficiently, leading to reduced computational cost during inference. Instead of building a separate set of parameters for every possible input pattern, the model can reconfigure its internal architecture on the fly, distributing specialized knowledge across layers as needed. This dynamic reuse of parameters can lead to more compact models that are still highly effective at capturing complex relationships in the data. Also the models ability to distribute knowledge across its layers can lead to better resisting to total-forgetting.

\subsection{Iterative and Recursive Computation}
The network operates in an iterative manner. After the first set of hidden states and routing decisions is made by each layer, one iteration completes, the information is passed through the network once again, that passage is considered the next iteration, and at each layer, new routing decisions are made. This process continues until the model reaches the limit in iterations (or output reaches the output layer, but we wanted to limit the model's processing time in order to gain more control during training, thus we use a maximum number of iterations). Unlike traditional models, where information only flows in one direction from layer to layer, our model’s computation can be seen as a form of iterative processing, where information is continually refined and re-routed based on the input.

\begin{figure}
    \centering
    \begin{subfigure}[t]{0.30\textwidth}
        \includegraphics[width=\linewidth]{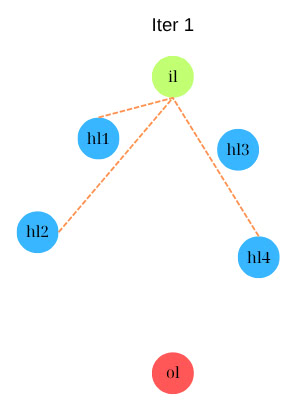}
        \caption{First iteration}
    \end{subfigure}
    \hfill
    \begin{subfigure}[t]{0.30\textwidth}
        \includegraphics[width=\linewidth]{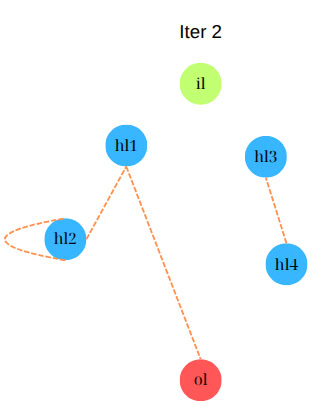}
        \caption{Second iteration}
    \end{subfigure}
    \hfill
    \begin{subfigure}[t]{0.30\textwidth}
        \includegraphics[width=\linewidth]{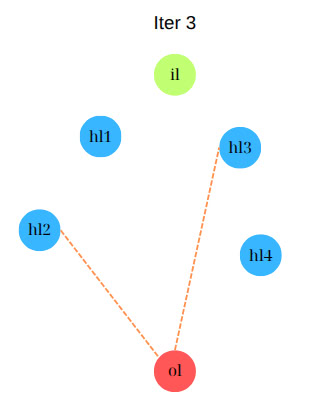}
        \caption{Third iteration}
    \end{subfigure}
    \caption{Iterative information flow of H-model.}
    \label{fig:h_model_on_iterations}
\end{figure}

This iterative aspect allows the network to perform more complex reasoning, with multiple layers having the opportunity to interact and refine the information being passed through them. In some cases, the routing could even be recursive, where information might flow through certain layers multiple times before it reaches the output. This flexibility not only increases the model’s capacity to handle more complex relationships but also introduces an additional level of adaptability, where the network can adjust its processing power depending on the complexity of the task at hand.

\subsection{Flexibility and Adaptation}
Ultimately, this model moves beyond the static nature of traditional neural networks and introduces a new paradigm where the model is constantly adapting itself based on the task and data it is faced with. Rather than requiring a fixed architecture that must be carefully tuned, the network adapts its structure on a per-input basis. This ability to dynamically adjust to different kinds of data, tasks, or even modalities makes the model more flexible and robust, opening new avenues for multimodal learning and complex, data-driven reasoning.

Through this adaptive design, the model effectively acts as a highly flexible system capable of tailoring its architecture to the needs of each individual input. This represents a significant departure from the conventional wisdom in deep learning, where the focus has often been on improving a fixed architecture's performance. In contrast, the proposed model shifts the focus toward dynamic, input-driven structural adaptation, allowing for a deeper, more modular understanding of the data it processes. Potentially in the margin of modern Large Language Models, H-model could evolve even to general-purpose models.

\begin{figure}[!h]
  \centering
  \begin{subfigure}[t]{0.5\textwidth}
        \includegraphics[width=\linewidth]{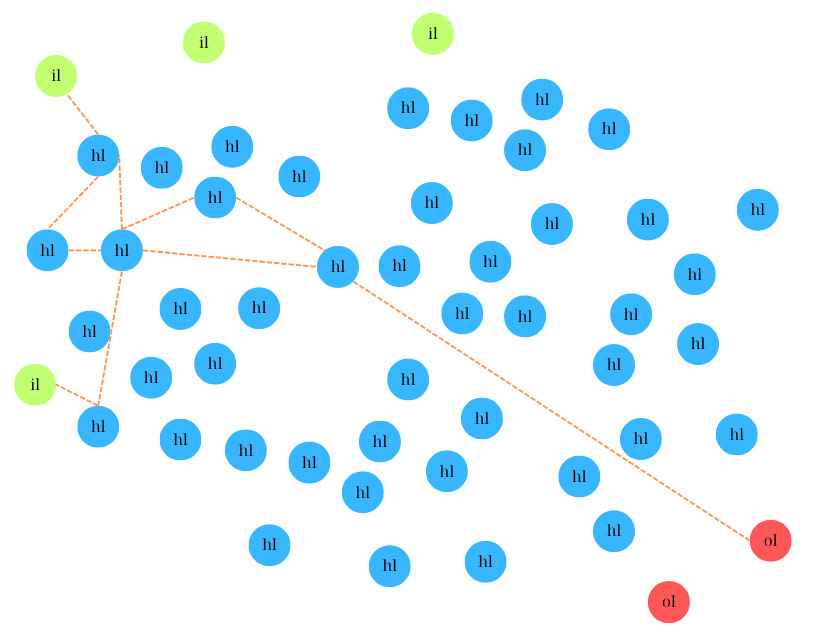}
        \caption{How the H-model would scale if its parameter magnitude were comparable to that of modern LLMs.}
  \end{subfigure}
  \label{fig:fig:h_model_scaleup}
\end{figure}
\section{H-Model}
\subsection{From the Concept to the Architecture}

The conceptual foundation of the H-Model lies in challenging the rigid, sequential layer execution of classical neural networks. Rather than statically channeling information from one fixed layer to the next, the H-Model introduces a dynamic routing mechanism that allows each layer to decide — based on the data — where its computed features should flow next.

To make this idea trainable and differentiable, we define each layer as emitting two outputs at every iteration: a hidden state $h_i \in \mathbb{R}^d$ and a connection signal $c_i \in [0,1]^L$, where $L$ is the number of hidden layers $H$ plus the number of output layers $O$. The connection signal is computed from the hidden state using a learned mechanism — not necessarily a single fixed transformation. For instance, in transformer-based layers, we extract the routing signal from a special embedding (e.g., the final token in the sequence), but in convolutional variants, it could come from a dedicated channel or learned head.

The resulting vector $c_i$ acts as a soft routing mask, controlling the flow of information from the current layer to other layers in the network. These weights are continuous, ensuring that the overall model remains fully differentiable:
\[
c_i = \text{RoutingHead}(h_i)
\]
where \texttt{RoutingHead} can be a task-specific or architecture-specific function learned jointly with the rest of the network.

Once computed, the routing weights guide how each layer distributes its hidden state to other layers of the H-Model. At each iteration, incoming contributions to a layer are aggregated (currently via simple addition):
\begin{equation}
\hat{h}_j^{(t)} = \sum_{i=1}^{H} c_{ij}^{(t)} \cdot h_i^{(t)}
\label{eq:aggregation}
\end{equation}
Where \( \mathrm{c}_{ij} \) denotes the connection weight to the hidden state from layer \( i \) to layer \( j \).

This aggregation respects gradient flow and encourages distributed, collaborative computation among layers. Routing is performed for a fixed number of iterations $T$, simulating depth through iterative refinement rather than architectural stacking. We believe that this aggregation of hidden states that share the same connection target by addition is an expedient but suboptimal implementation, as it does not explicitly contain the information of the source of the hidden state. So this way of passage can be reconsidered.

To ensure stability over repeated passes, we apply normalization after each aggregation step, scaling the resulting hidden state by the number of contributors. Moreover, layers can route information to themselves, enabling optional state persistence.

After $T$ iterations, all states directed to the output layer are summed over $T$ and processed to yield the final result. A final normalization stage (across iterations) ensures that the accumulated signal remains stable and interpretable.

In effect, the H-Model introduces a differentiable meta-structure over classical layers, wrapping them in a dynamic, data-driven computation graph. This architecture moves toward a more modular, flexible model that can adapt its internal flow to the task and data at hand — even using existing components like transformers in a novel, structural context.

\begin{figure}[h!]
    \centering
    \begin{subfigure}[t]{0.4\textwidth}
        \includegraphics[width=\linewidth]{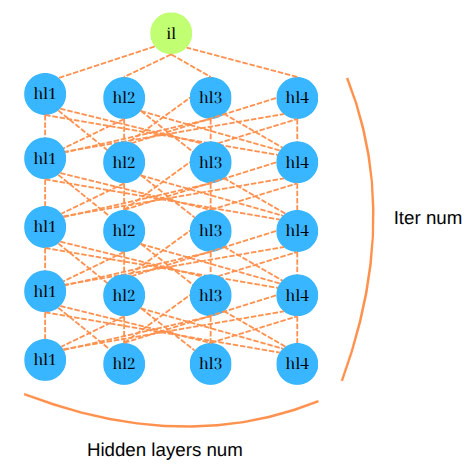}
        \caption{Fully structured and well-defined H-model visualization.}
    \end{subfigure}
    \hfill
    \begin{subfigure}[t]{0.3\textwidth}
        \includegraphics[width=\linewidth]{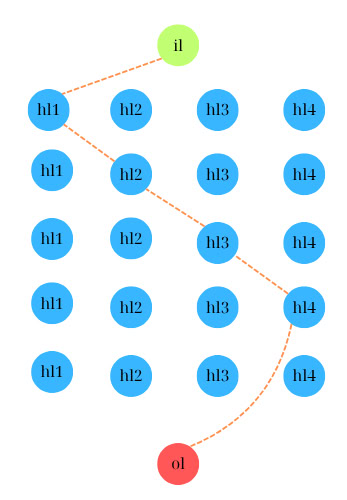}
        \caption{Traditional forward propagation as a subset of possible propagations and routing decisions of our model.}
    \end{subfigure}
    \label{fig:h_model_structured}
\end{figure}

\subsection{Training Constraints and Regularization}

The flexibility of the H-Model's routing mechanism introduces constraints that are essential to maintain coherence and trainability. These constraints ensure that the dynamic connections between layers can function seamlessly while preserving gradient flow and model stability.

\paragraph{Uniform Hidden State Dimensionality.}
At the core of the architecture is the idea that any layer can route its hidden state to any other layer. To make this viable, we enforce a strict constraint: \textbf{all hidden states across layers must share the same dimensionality}. That is, for all layers $i$, the hidden representation $h_i \in \mathbb{R}^d$ must lie in the same latent space. This allows for direct aggregation of incoming states, without any projection or alignment overhead.

\paragraph{Soft Routing for Differentiability.}
To enable training via backpropagation, all routing weights $c_{ij} \in [0,1]$ are computed via a differentiable activation function. We primarily use the sigmoid function due to its interpretability and simplicity, although alternatives such as softmax or Gumbel-softmax can be explored for sharper or more selective routing.

\paragraph{Aggregation Stability.}
When a target layer receives signals from multiple sources, their hidden states are aggregated. Our default aggregation strategy is simple summation \ref{eq:aggregation}.

This aggregation is followed by normalization to prevent uncontrolled growth in representation magnitude — especially as the number of contributors or iterations increases. In particular, we normalize the aggregated hidden state by the number of contributing layers and later normalize again when passing the final signals to the output layer after $T$ iterations.

\paragraph{Gradient and Flow Control.}
We limit the number of iterations $T$ to avoid excessive depth and ensure tractable computation. Moreover, layers are permitted to route information to themselves, enabling a soft form of recurrence and memory, but this too is regularized through normalization to avoid state explosion.

These constraints strike a balance between architectural flexibility and training feasibility, enabling the model to evolve computation paths while remaining grounded in practical deep learning principles.

\paragraph{Designated Input and Output Layers}

To maintain architectural clarity and tractability, the model distinguishes between standard hidden layers and special-purpose input and output layers. These designated layers are restricted in their behavior: the \textbf{input layers} are responsible solely for processing raw inputs and emitting initial hidden states during the first iteration, while the \textbf{output layers} serve exclusively to process the accumulated hidden states and produce final predictions at the end of the computation loop.

This restriction ensures a clean separation of roles, preserving the interpretation of the model's hidden state evolution as a computation graph flowing from input to output. Importantly, the model may employ \emph{multiple} such layers to handle various input modalities (separate layers for textual and visual input) or to support different output heads (multi-task learning scenarios). These layers do not participate in the iterative message-passing process outside of their defined roles.

\subsection{Overall Architecture}

The proposed architecture operates as a \textit{meta-model}, abstracting away from specific architectural details and focusing instead on the dynamic routing of information. The minimal structural unit of the model is what we refer to as a \textbf{layer}, though each such layer may internally consist of a complete deep network (e.g., a transformer block or a stack of convolutional layers). But we do not recommend to use deep networks as your layer, because it limits the ability of the model to actually decide the routing pathes. This design allows the architecture to retain flexibility while maintaining consistent interfaces between layers.

The model is organized into three distinct sets of layers:
\begin{itemize}
  \item \textbf{Input layers}, which receive and encode the raw data;
  \item \textbf{Hidden layers}, which perform iterative computation and routing;
  \item \textbf{Output layers}, which aggregate and decode the final hidden states.
\end{itemize}

Any architecture that satisfies the hidden state compatibility constraint (i.e., consistent dimensionality across layers) can be integrated. This makes the framework highly general and extensible to diverse tasks and data types. We believe that the model's full potential may be realized by integrating as diverse a set of models as possible—each capturing the input through distinct functional mechanisms. This diversity equips the H-model, as we call it, with a broader computational arsenal.

The architectural novelty and computational potential of the model emerge most clearly in its forward pass, which is discussed in detail in the next subsection.

\subsection{Forward Pass and Iterative Computation}

The defining characteristic of the H-model's forward pass is its departure from traditional feed-forward computation. Rather than statically passing input through a fixed sequence of layers, our model unfolds its computation over a series of $T$ iterations, during which intermediate representations are dynamically routed between hidden layers. This approach significantly increases the model's ability to effectively use its parameters.

While the input and output layers are invoked only once per forward pass, the hidden layers are activated across $T$ recurrent iterations. As a result, the effective compute time grows linearly with $T$, making inference approximately $T$ times more expensive than a comparable feed-forward model of the same parameter count. However, this trade-off provides substantially more expressive routing behavior and the ability to specialize computations across time and topology. We deliberately avoided vectorized parallelization across hidden layers, as it would restrict the model to specific layer architectures. However, there is an approach that would help us, as the inventors of it faced with the same problem - Sparse decisions for MoE \cite{sparse_adaptive_moe}. We emphasize that the modular structure of our model naturally lends itself to distributed computation, making it well-suited for scalable, parallel environments. 

We also considered as architectural improvements for the model for the future scaling - using special windowed connections, where the layers can pass hidden states only to the nearest layers, kinda like Swin \cite{swin} processes the input image. This will also allow for easier models extension. Even after training you would be able to add layers to the model and just connect them to the edge of the network. 

Similar approach of this modality was considered due to its easy implementation. As our H-model is technically a wrapper for other models, we could use several H-models as layers, its own minimal block. This extension of the model could go infinetelly opening us a new axis of modality.

\paragraph{Forward Pass Logic.}
Each layer, during iteration $t$, receives a weighted combination of hidden states from other layers based on learned routing weights. These weights are computed from the layer’s current hidden state and normalized across all possible destinations using a sigmoid activation. Routing weights guide how strongly a layer’s output contributes to each potential receiver. The output layer only receives signals in the final step, after $T$ iterations, where it aggregates and processes all accumulated information that was directed to it.

\vspace{1em}
\noindent The following pseudocode summarizes the forward pass process:

\begin{lstlisting}[language=Python, caption={Forward Pass in H-model}, label={lst:forward_pass}]
# Deff: x - list of unique raw input, T - number of iterations, H, O, and I - are numbers of hidden output and input layers respectively, i_blocks, h_blocks, and o_blocks - are respectively sets of input, hidden, and output layers.

# Step 1: Process input
state = 0
outs = 0

for i in range(I):
    substate, con = i_blocks[i](x[i])
    state = state + con * substate
    # (batch, O + H, ...)
state = norm_by_contributors(state)
outs = outs + state[:, -O:]
# last hidden states passed to the output
# Step 2: Iterative processing
for j in range(T):
    new_state = 0
    for i in range(H):
        substate, con = h_blocks[i](state[:, i])
        substate = substate + IterEmb(j)
        new_state = new_state + con * substate
    state = norm_by_contributors(new_state)
    outs = outs + state[:, -O:]
outs = outs / max(T, 1) # norm by T
# Step 3: Outputs processing
final_out = []
for i in range(O):
    substate, con = o_blocks[i](outs[:, i])
    final_out.append(substate)

return final_out
\end{lstlisting}

During each iteration, a given hidden layer may receive inputs from multiple upstream layers. These inputs are weighted by the connection coefficients computed via the routing mechanism. However, to prevent the hidden state's magnitude from unintentionally growing with the number of contributors, we introduce a normalization mechanism that scales the aggregated hidden state by the \textit{sum of incoming connection weights or number of contributors}---but only when this sum exceeds 1. To allow this to happen, we also have to save all the connections during each iteration temporarily.

We define the aggregated pre-normalized state as in \ref{eq:aggregation} and the norm as :

\begin{equation}
\hat{h}_j^{(t_n)} = NormByContributors(\hat{h}_j^{(t)})
\label{eq:norm_by_contributors_formal}
\end{equation}

Which is the same as:

\begin{equation}
\hat{h}_j^{(t_n)} = \frac{\hat{h}_j^{(t)}}{\max\left(1, \sum_{i=1}^{L} c_{ij}^{(t)} \right)}
\label{eq:norm_by_contributors}
\end{equation}

This method ensures that the hidden state at each layer does not get arbitrarily amplified when many strong connections are directed toward it. $t_n$ means that the hidden state is still the output of the same iteration, but already normalized and ready as input to the next iteration.

\subsection{Individual layers}

To easily follow the constrains, we also define architectural units or layers as a simple model with a wrapper, so-called $WithConnection$. This allows us to add a unique routing mechanism to any model based on its properties.

\vspace{1em}

\begin{lstlisting}[language=Python, caption={WithConnection torch implementation}, label={lst:with_connection_wrapper}]
class WithConnection(nn.Module):
    def __init__(self, model, con_number):
        super().__init__()
        self.model = model
        self.con_number = con_number

    def forward(self, x):
        out = self.model(x)
        con = RoutingHead(out)

        return out, con
\end{lstlisting}

Where, as was mentioned earlier, $RoutingHead$ is any function that processes the layer`s output to return connection weights. We usually used a simple FF model with sigmoid activation on top.

\paragraph{Sharpening}

We usually formed the $RoutingHead$ with a sharpening mechanism applied to the sigmoid. This is similar to the Gumbel-softmax or DINO's \cite{dino, dinov2} use of temperature, but as was mentioned earlier, we used a sigmoid for interpretable visualization of connections during the forward pass, some of which you can see in the Experiments section. This can be easily described as:

\begin{equation}
con = sigmoid(\frac{conlogits}{\alpha})
\label{eq:sharp_routing}
\end{equation}

With this logic, changing the parameter $\alpha$ to a very small value would result in more sparse connections imitating the behavior of large pretrained models, making the model more confident in decisions. And thus changing $\alpha$ to a very large value would result in more dense connections, encouraging the model to use as many of its parameters as possible.

Knowing that, we also decided to add one more auxiliary parameter on the H-Model level $\alpha_{rate}$. That parameter would decide on the change of the parameter $\alpha$ on each iteration, by multiplication. This way we could make the model create more dense or sparse connections with iterations. Also, we decided to leave these parameters as hyper-parameters, as we can change them at any time during training or inference.

Optionally, we can apply \textit{discretization at inference time} by taking the top-\( k \) most active routes, significantly reducing the computational overhead of the full forward pass:

\begin{equation}
\tilde{\mathbf{c}}_i^{(t)} \leftarrow \text{TopK}\left(\mathbf{c}_i^{(t)}\right)
\label{eq:inference_con}
\end{equation}

This enables the model to operate through a \textit{sparser computation graph} post-training, enhancing inference efficiency without requiring retraining.
\section{Separable Depth and Layers}

In most conventional neural network architectures, the number of layers directly corresponds to the model's depth: each layer represents a fixed stage in a sequential computation graph. Those terms are quite interchangeable nowadays. However, in the proposed H-model, this assumption is relaxed. The notions of \textit{layer count} and \textit{computation depth} are treated as independent hyperparameters, leading to a more modular and flexible structure.

\subsection{Depth as Temporal Computation}

 Rather than stacking layers sequentially as in a traditional feed-forward network, we allow these $H$ modules to communicate over $T$ iterations, forming a recurrent and dynamic information flow. This transforms depth from a spatial property (number of layers stacked) into a temporal one (number of steps across a dynamic topology or number of iterations T).

\subsection{Layer as Functional and Memory unit}

Each layer in the H-model represents a functional unit capable of processing a hidden state and emitting routing signals. These layers are independent in parameters and thus can specialize in different transformations or modalities. By increasing $H$, the model expands its representational and specialization capacity, much like increasing the number of neurons.

\subsection{Design Trade-offs}

This decoupling of depth and layer count introduces a spectrum of design choices. Consider the following contrasting setups:

\paragraph{High Layer Count, Low Iterations.}  
A configuration with $H = 10$, $T = 2$ prioritizes structural diversity. The model can activate and combine a large number of modules, encouraging parallel and distributed representations. It is particularly suitable for multi-modal inputs or tasks requiring compositional reuse of domain-specific submodules. However, since the number of iterations is low, the effective depth of reasoning is limited. Thus, this setup is suitable for the benchmarks with high data diversity but low data complexity.

\paragraph{Low Layer Count, High Iterations.}  
A configuration with $H = 2$, $T = 10$ emphasizes computation depth. The same few layers are reused across multiple time steps, allowing deep feature transformations with a limited parameter budget. This favors compact models that require strong compositional reasoning or abstraction, similar to recursive processing in symbolic tasks. Such a setup encourages the model to reuse its layers, resulting in more effective parameter utilization. This setting of reuse of the same layers was introduced to us many times in the modern science community and has proven to be applicable.

\paragraph{Balanced Configuration.}  
Intermediate setups (e.g., $H = 4$, $T = 4$ or $H = 10$, $T = 10$) offer a trade-off between memory and depth. These can serve as general-purpose configurations when the task structure is not fully known in advance. By visualizing the model's connection during inference, we could gather insights on which possible setups are most suitable for the task.

\subsection{Biological Analogy}

Interestingly, this separation mirrors certain properties of biological systems. In the brain, individual regions (analogous to layers) may remain structurally static while their interactions and firing patterns evolve over time. Learning occurs not only by changing the modules (synaptic weights), but also by refining the timing and routing of information between them. The H-model inherits this view, enabling the reuse and dynamic coordination of computational units. Allowing only a part of the network be fired to deliver the output result gives the model ability to distribute its knowledge in different routing patterns between its layers, similarly as only some parts of human brains are activated with different inputs and how parts of brains directly specialize in different modalities.

\subsection{Implications for Learning and Flexibility}

Separating $H$ and $T$ introduces valuable flexibility into the architecture design. Unlike traditional models where increasing depth inherently means increasing parameter count, here we can independently tune depth (temporal iterations) and width (layer variety), enabling fine-grained architectural control.

\begin{itemize}
    \item \textbf{Increased Parameter Efficiency:} The model can simulate greater computational depth through iterations without adding new parameters. A smaller number of highly reused layers (high $T$, low $H$) can act like a deep model with a small footprint.
    
    \item \textbf{Improved Specialization and Modularity:} A larger number of diverse layers (high $H$, low $T$) allows the model to distribute knowledge across specialized submodules. This improves interpretability, reusability, and multi-domain capability.
    
    \item \textbf{Flexible Scaling:} Since the computational graph is formed dynamically at inference, we can scale the model along either axis ($H$ or $T$) to meet specific deployment constraints—such as latency or memory limits.
    
    \item \textbf{A New Axis of Flexibility:} Most importantly, the decoupling introduces an additional axis of model design. Traditional models vary only along depth and width (stacked layers or increased feature dimensions), but the H-model can vary in its \textit{layers permutations and utilization}—its ability to reconfigure computation paths at runtime. This makes the architecture inherently more adaptive and opens up a broader space for future innovations. Instead of choosing a model architecture from one linear combination of such parameters as depth and number of layers (1 to 1), we break this limitation and allow you to chose hyper-parameters from the full 2D space.
\end{itemize}

To recap, this flexibility is not just a minor architectural detail, it reflects a shift in how we can think about model capacity and structure. By decoupling depth from the number of layers, the H-model enables a richer space of design choices, where computation, memory, and specialization can be independently controlled. This opens the door to architectures that are not only more adaptive to diverse tasks, but also more aligned with the evolving demands of efficient and modular machine learning systems.

\section{Ensembles Perspective}

In this section, I want to highlight a particularly exciting aspect of the H-model: its ability to integrate multiple state-of-the-art (SOTA) models into a single metamodel at the layer level. This means we can literally combine different models—each potentially trained on different tasks or modalities—into one cohesive architecture. By doing so, we aim to harness their individual strengths and knowledge, effectively creating a more powerful and versatile system.

The key here is the H-model's modular design and dynamic routing capabilities. Each layer can be thought of as a self-contained module, possibly representing a layer of a different SOTA model. Through learned routing mechanisms, the model can determine the optimal pathways for information flow, allowing it to leverage the most relevant modules for a given task. This setup not only facilitates knowledge sharing across layers but also enables the model to adaptively specialize its computations based on the input.

This approach aligns well with the current trend towards modular models, especially in multimodal tasks like image-to-text, video-to-text, and visual question answering (VQA). Traditional models often require separate architectures or pathways for each modality, leading to redundancy and inefficiency. In contrast, the H-model's shared hidden layer space allows for more effective integration of multimodal inputs. Instead of segregating modalities, the model can process them within a unified framework, promoting better interaction and understanding between different types of data.

Moreover, this design offers practical benefits. For instance, we can extend the model by adding new layers (modules) without retraining the entire system. These new modules can be connected to the existing network, and through the routing mechanism, the model can learn how to incorporate them effectively. This flexibility is particularly valuable in rapidly evolving fields where new models and modalities are continually emerging. The only drawback here is that the model still has to learn connections and routing mechanisms to effectively model routing decisions.

In a way, our model already mimics and generalizes some ensemble learning strategies. Even though it’s not a classic ensemble by training multiple models separately, its structure and dynamics often resemble what real ensembles are doing — and sometimes go even further.

Let’s start with how the layers work in parallel. In every iteration of the forward pass, each hidden layer processes the received hidden state independently and contributes its output to other layers. This is very similar to how traditional ensemble methods operate — where multiple models process the same or similar inputs and combine their predictions. The closest analogy in modern deep learning might be the Mixture of Experts (MoE) models \cite{adaptive_moe, sparse_adaptive_moe}, where only a subset of experts is selected for each input. My model does something similar — the routing mechanism learns which layers to activate softly, with the possibility of sharpening them into harder decisions if needed.

Then there is the depth of the model. Since the hidden layers are reused across $T$ iterations, we get something similar to deep recurrence. This part aligns quite well with how classic deep networks \cite{backpropagatingerrors} function — adding more layers to increase representational capacity. But the H-model separates depth from parameter count. So instead of just stacking layers to increase depth, we iterate over a flexible routing topology. You can think of it like deep learning’s depth is still there, just distributed over time, not layers.

Skip connections? That’s in there, too. Because any layer can route its hidden state directly to the output layer, we technically allow behaviors similar to ResNet-style \cite{resnet} shortcuts. It’s even more flexible in our case, because those connections can be learned dynamically based on input, not statically hardcoded into the network’s shape.

Another important parallel is with neural architecture search (NAS) \cite{zoph2017neuralarchitecturesearchreinforcement, pham2018efficientneuralarchitecturesearch}. In NAS, researchers or algorithms search for the best topology for a specific task. But in the H-model, this search is embedded directly into training. The routing mechanism learns the best way to use the available layers for each kind of input. It’s like continuous NAS — but during forward pass, not beforehand.

All of this combines beautifully with the previously discussed ability to insert diverse architectures into the model. You can throw in a ViT, a CNN block, a language model transformer, and maybe even a diffusion decoder — and let them co-exist in one routing space. The model can figure out when and how to use them. That’s real ensemble-level behavior but tightly integrated inside a single forward graph.

And let’s not forget the real benefit: this ensemble-like setup doesn’t just increase performance, it also enables better modularization and transfer. Because the hidden state lives in a shared space, information flows freely between parts of the model, no matter what architecture those parts are using. That’s especially useful for multimodal tasks, where we don’t want models for image, text, and video to live in isolated silos — we want them to collaborate.

In short, the H-model doesn’t just look like an ensemble — it’s a generalization of one. It combines the strengths of ensemble learning, routing mechanisms, dynamic depth, and even skip architectures like ResNet, wrapping them into one coherent and trainable system. All while keeping the door open for modularity, extensibility, and creative combinations that typical static models just can't do. And that all is achieved by a simple idea. 

Interestingly, this approach also echoes some of the early motivations behind the Inception \cite{inception} architecture. Inception networks were designed to allow different types of computations—1x1, 3x3, 5x5 convolutions, and pooling—at the same layer, letting the model decide which scale of information was most useful. Our model generalizes this idea one step further: instead of just mixing spatial filters, we allow the routing system to dynamically pick across entire blocks of computation, which could be anything—from a transformer to a convolution or even a pretrained module. In that sense, our routing mechanism acts like a learned, dynamic version of the Inception module, but stretched across both depth and architectural diversity.

\subsection{The Closest Analogy: A Generalization of MoE \cite{adaptive_moe, sparse_adaptive_moe} and Attention \cite{attention, vaswani2023attentionneed}}

In many ways, our model can be seen as a generalization of both attention mechanisms \cite{attention} and Mixture of Experts (MoE) \cite{adaptive_moe, sparse_adaptive_moe}. It borrows critical ideas from both but goes significantly further by operating at a more flexible and abstract level. If attention mechanisms compute "what to attend to" among tokens, and MoE chooses "which experts to activate" for each input, then our model decides "how to dynamically route computation between layers" — at every step, for every input, across time.

\paragraph{Similarity with MoE \cite{adaptive_moe, sparse_adaptive_moe}:}

Let’s first look at MoE \cite{adaptive_moe, sparse_adaptive_moe}. MoE \cite{adaptive_moe, sparse_adaptive_moe} models typically follow a one-to-many pattern: one input is fed into a set of experts, and the gating mechanism softly or sparsely selects a small number of experts to process it. These are then aggregated and passed forward. In our case, we also have selective activation, but we allow each layer to selectively send information to any number of other layers — and receive from any number, too. So instead of one-to-many (input to experts), we enable a many-to-many connection scheme. Every layer can become both an "expert" and a "client" dynamically, depending on what the input and routing mechanism decide.

Another key difference is that MoE \cite{adaptive_moe, sparse_adaptive_moe} typically only performs expert selection once, at a specific depth in the model — a fixed architecture location. Our model enables routing decisions at every iteration, across all hidden layers, making it not just adaptive in width (which experts to use), but also in depth and time. You can think of this as continuous, learned dynamic architecture that evolves during the forward pass.

Also important is that MoE \cite{adaptive_moe, sparse_adaptive_moe} usually treats its experts as homogeneous blocks, often sharing architecture or task design. But we don’t make that assumption. In fact, we explicitly embrace architectural diversity — one layer might be a transformer block, another a convolutional module, and another a pretrained vision encoder. Routing across these becomes not just about selecting the best expert but also about combining fundamentally different forms of computation and representation in a task-driven way.

So while the behavior resembles MoE \cite{adaptive_moe, sparse_adaptive_moe} at a high level — selective routing, sparse activation, specialization — the H-model does it in a richer, more general setting, with temporal depth, topology learning, and architectural heterogeneity.

\paragraph{Similarity with attention \cite{attention, vaswani2023attentionneed}:}
On the attention side \cite{attention}, we also share similarities. At each step, we learn soft assignments of where information should flow — which is conceptually close to attention scores. In fact, if we visualize the routing coefficients at each step, we can think of them as topological attention masks, showing how computation is distributed across the network graph. But unlike attention, which usually operates between tokens based on feature similarity, our routing operates between layers — and is guided by both the input and the current global state of the network.

It’s also worth noting that attention tells you how much of one token should contribute to the current one. Our model’s routing tells you how much of one layer’s output should be sent to another layer’s input in the next iteration. So while the logic is analogous, the granularity and target of computation differ significantly.

If anything, the closest cousin to our mechanism might be sparse MoE \cite{adaptive_moe, sparse_adaptive_moe} models with dynamic sparsity and top-down control — but allowing dynamic depth, many-to-many connections and different modalities by operation on layer level.

\section{Experiments}

\subsection{Purpose and Goals}

The primary purpose of this article is to \textbf{investigate the capabilities of the H-model} and explore its ability to exhibit flexible architectural behaviors through its dynamic routing mechanism. As a preliminary study, the goal is not to achieve state-of-the-art performance but to demonstrate the feasibility of the H-model’s core capabilities and validate some key architectural hypotheses.

The first goal of this work was to \textbf{prove that the H-model can generate constant architectures} by creating stable routing paths. By routing through layers in a fixed, repeatable manner, the model can simulate traditional architectures that process data in a known sequence. This is essential for understanding whether the model’s routing mechanism can maintain structural consistency.

Secondly, we aimed to \textbf{validate the model’s ability to select different architectures based on input data}, enabling the model to dynamically choose its routing paths. This behavior allows the model to adjust to the specific needs of the data it processes, effectively creating context-dependent architectures. The hypothesis was that different inputs would result in distinct paths through the network, thus enabling the model to respond flexibly to varying tasks.

Another key goal was to \textbf{demonstrate the model’s ability to combine pretrained models}, leveraging the layers of existing architectures. This would enable the H-model to build on the knowledge of established networks, providing an easy mechanism to \textbf{incorporate external, pretrained models} into its framework. Specifically, the model’s routing system would determine how layers from different pretrained models could be connected and utilized.

Finally, we aimed to test whether the \textbf{H-model can process different input types using distinct input layers and produce varying results based on these inputs through output layers}. This experiment was designed to assess the model’s ability to dynamically route information across different layers depending on the nature of the input, a key aspect of its flexibility in handling diverse tasks. We aim to test the models ability to distribute knowledge, particularly we wanted to answer the question whether the model would create an architecture where it is divided into some parts of in-domain knowledge (like part that is responsible only for image processing or for language) or highly share knowledge from different domains without expert subdivision.

This study serves as a preliminary investigation into the potential of the H-model to handle modular, dynamic, and input-dependent routing while incorporating pretrained models and offering a new way to explore model composition.

\subsection{Experimental Setup}

Given the resource constraints of this study, in most cases, we limited our evaluation to unimodal settings—specifically, natural language tasks. Although the H-model is designed with multimodal flexibility in mind, validating its superiority in that context would require significantly larger experimental infrastructure and datasets. As such, we focused on the language domain, where architectures and benchmarks are mature and well-understood.

Throughout our experiments, we used the Transformer layer as the core building block of the H-model. Each “layer” in the H-model corresponds to a single Transformer block, allowing us to isolate and observe the effects of dynamic routing and iteration. This choice was motivated by the Transformer’s ubiquity in NLP and its well-documented performance characteristics, which provide a solid baseline for comparison.

To evaluate the behavior of the H-model, we compared it against standard Transformer-based architectures with approximately the same parameter count. This allows us to analyze the trade-offs between dynamic routing and static depth in comparable settings. In both cases, models were trained from scratch using similar optimization parameters, ensuring a fair evaluation of architectural behavior.

\subsection{Stable Architectures}

One of the most consistent observations from our experiments was that the H-model is fully capable of converging to a stable, task-specific architecture by learning constant routing paths across all inputs. In many cases, these paths reflect efficient computation strategies for the given task and remain consistent throughout training and evaluation. This supports the idea that the model can act not only as a dynamic routing framework but also as a neural architecture search mechanism.

\begin{figure}
    \centering
    \includegraphics[width=0.8\linewidth]{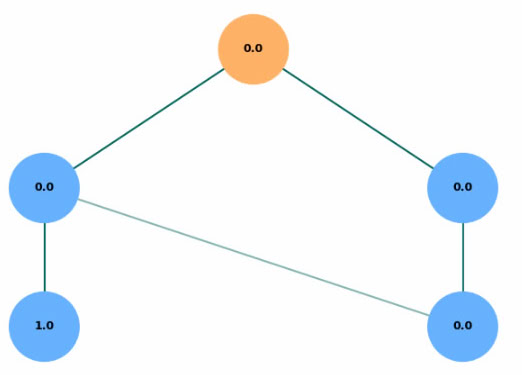}
    \caption{An example of trained routings of the model on a dataset that requires simple reasoning. This architecture is stable to different inputs and is a good example of a stable architecture. In this case, it is easy to see that this model actually imitates the traditional forward pass, but we have seen a lot of cases of more complex routings that cannot be reproduced with traditional models.}
    \label{fig:stable_architecture_proofwritter}
\end{figure}

Importantly, this behavior opens the door to a wide range of studies on emergent architectures—how they form, when they stabilize, and how they correlate with downstream performance. While this work only scratches the surface, we provide visualizations of these converged routing paths \ref{fig:stable_architecture_proofwritter}.

On the current task we also decided to make a comparison with traditional models and H-model's hyper-parameters. The dataset we used is \textbf{ProofWriter} \cite{proofwritter}, which provides a structured logical reasoning task that is perfect for observing how information flows through the model. It includes varying levels of complexity and multi-step reasoning, making it a solid choice for evaluating dynamic routing behaviors.

\begin{figure}[!h]
    \centering
    \begin{subfigure}[t]{0.45\textwidth}
        \includegraphics[width=\linewidth]{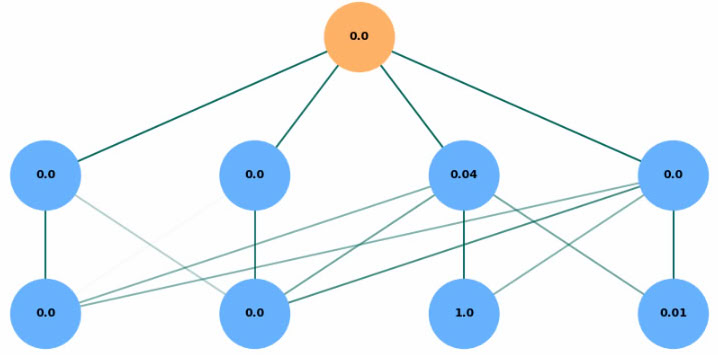}
        \caption{First pattern of routing paths.}
    \end{subfigure}
    \hfill
    \begin{subfigure}[t]{0.45\textwidth}
        \includegraphics[width=\linewidth]{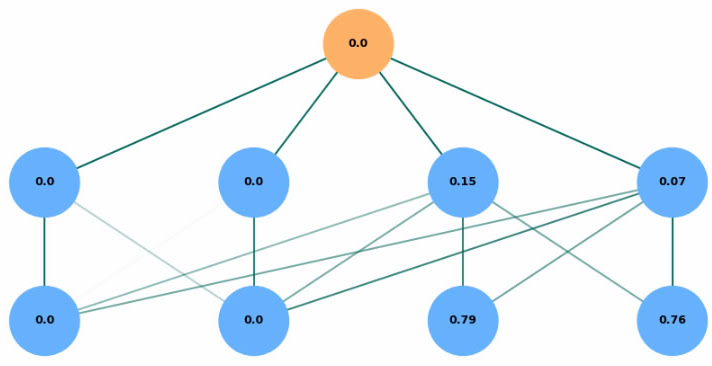}
        \caption{Second pattern or routing paths.}
    \end{subfigure}
    \caption{An example of trained routings of the model on a dataset that requires simple reasoning. This model trained two main routing patterns that are based on the inputs.}
    \label{fig:partially_stable_architecture_proofwritter}
\end{figure}

Also in the same setting and with the same dataset, results of a partially stable architecture with two main patterns of behavior were achieved \ref{fig:partially_stable_architecture_proofwritter}.

\begin{figure}[!h]
    \centering
    \begin{subfigure}[t]{0.45\textwidth}
        \includegraphics[width=\linewidth]{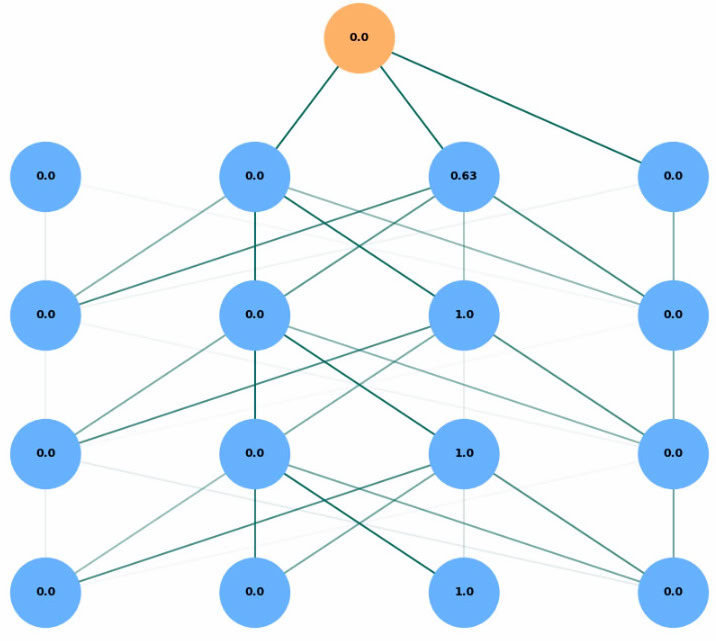}
        \caption{First pattern or routing paths.}
    \end{subfigure}
    \hfill
    \begin{subfigure}[t]{0.45\textwidth}
        \includegraphics[width=\linewidth]{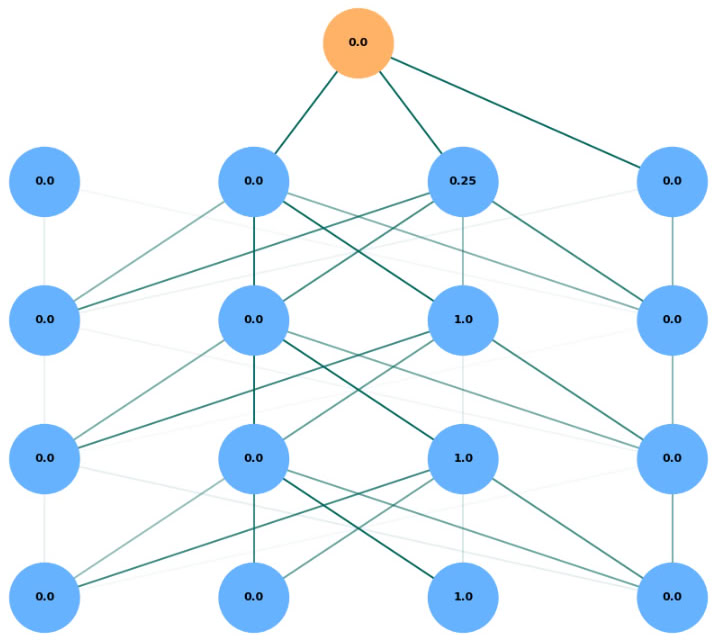}
        \caption{Second pattern or routing paths.}
    \end{subfigure}
    \caption{An example of a deeper architecture routing. Though at the moment this experiment does not focus on the adaptive architectures that change based on the input, we can already observe such a behaviour, but only in weak differences in possible routing paths.}
    \label{fig:partially_stable_architecture_proofwritter_deep}
\end{figure}

To further demonstrate the varying behavior of the model, we included an example of a more complex routing path in Figure \ref{fig:partially_stable_architecture_proofwritter_deep}, where the model learns to utilize deeper iterations and non-trivial routing jumps between layers. This behavior would be very difficult to simulate in static architectures, highlighting one of the strengths of our framework.

\subsection{Quantitative Comparison}

To complement our qualitative findings, we also ran a set of controlled experiments to compare the H-model with traditional Transformer baselines. All models were trained on the ProofWriter dataset using a simple language modeling objective. Due to limited compute, most of the training runs were conducted for only 5 epochs, which is insufficient for full convergence. However, two of the smallest models ($T_{uc}$ and $H_{uc}$) were trained until convergence and achieved near-perfect accuracy ($> 0.999$), suggesting that more comprehensive experiments and higher-capacity datasets are necessary to fully evaluate the H-model’s strengths.

In these experiments, we also kept the model capacity intentionally low by using a hidden size of 128 and feed-forward size of 256. This constraint was important to ensure that performance differences arise primarily from routing and architectural choices rather than raw parameter count. \ref{tab:proofwriter_comparison}

\begin{table}[!h]
    \centering
    \begin{tabular}{|l|c|c|c|c|c|c|}
        \hline
        \textbf{Model} & \textbf{Layers} & \textbf{Iter} & $\alpha$ & $\alpha_{rate}$ & \textbf{ACC} & \textbf{Loss} \\
        \hline
        \multicolumn{7}{|c|}{\textbf{Traditional Model}} \\
        \hline
        ${Tr}_{uc}$ & 4 & - & - & - & 0.9998 & 0.001 \\
        ${Tr}_5$  & 6 & - & - & - & 0.77 & 0.56 \\
        \hline
        \multicolumn{7}{|c|}{\textbf{H-Model}} \\
        \hline
        ${Hm}_{uc}$ & 4(2) & 2 & 1 & 1 & 0.9998 & 0.001 \\
        ${Hm}_5$  & 6(4) & 2 & 1 & 1 & 0.73 & 0.68 \\
        ${Hm}_5$  & 6(4) & 4 & 1 & 1 & \textbf{0.79} & \textbf{0.51} \\
        ${Hm}_5$ & 6(4) & 10 & 1 & 1 & 0.72 & 0.67 \\
        \hline
    \end{tabular}
    \caption{Comparison of traditional and H-model configurations on ProofWriter task. The number of layers listed for H-models includes both active and total layers, shown as $L(\text{active})$. All models share the same embedding and feed-forward size.}
    \label{tab:proofwriter_comparison}
\end{table}

We observe that increasing the number of iterations $T$ leads to a trade-off between model convergence speed and full capasity. For instance, in the H-model with $T=4$, the loss decreased notably compared to $T=2$, indicating deeper and more refined computation paths that were possible to achieve in 5 epochs. However, a further increase to $T=10$ led to slight degradation in both accuracy and loss, likely due to the need in more time to converge.

\subsection{Routing and Stability in Deeper Architectures}

As noted earlier, the routing mechanism in the H-model shares a conceptual resemblance to residual networks (ResNets) in that it allows information to skip intermediate layers and reach deeper ones. This similarity becomes even more pronounced given our use of Transformer layers as the minimal building block, which inherently include residual connections. However, the H-model's dynamic routing introduces a critical difference: the ability to completely suppress or bypass a layer through a learnable routing weight that can be driven to zero. In contrast, ResNet-style skip connections always add the previous state to the current layer output, meaning the network must either learn to output near-zero residuals or carefully counteract prior activations. This architectural rigidity can limit ResNet’s stability in very deep configurations.

In our experiments, we consistently observed that H-models with increased iteration depth $T$ still converged—sometimes requiring more time, but always eventually stabilizing. This contrasts with traditional deep networks, where deeper architectures often experience degradation in trainability or convergence. We hypothesize that the greater stability of the H-model is due to its flexible, trainable routing: it can entirely skip redundant computation paths when necessary.

Furthermore, the interpretable visualizations of routing paths (e.g., Figure~\ref{fig:partially_stable_architecture_proofwritter_deep}) show precisely where the model decides to halt deeper computation. This behavior is emergent and varies across tasks and input patterns, offering unique transparency into model dynamics. These patterns suggest that the model self-regularizes its depth per input rather than following a fixed forward path. This could serve as a more efficient solution to the vanishing gradient and overfitting issues typically associated with deep architectures.

We believe this difference in skip connection semantics—where H-model routing acts as a selective gate rather than a forced additive path—plays a fundamental role in this stability. Unlike ResNet, which must always propagate some form of state forward, the H-model can opt out completely. This architectural distinction offers a simpler and potentially more robust mechanism for depth scaling and may explain the smoother training behavior we observed in deeper iterative variants.

By ignoring the deeper layers, our model can essentially just simulate models with fewer iterations or depth, thus reducing the problem of vanishing gradients in traditional approaches in some way.

\subsection{Adaptive Architectures: Multilingual Routing}

Following our investigation into stable architectures, the next natural question to ask was whether the H-model is capable of adapting its internal architecture dynamically depending on the input. To explore this behaviour, we designed a multilingual experiment using a language modeling task across six different languages. The chosen languages were selected to be as diverse as possible in terms of syntax, morphology, and language family, ensuring that the model would encounter distinctly different input distributions. To be precise we decided on the following list of languages:

\begin{enumerate}
    \item German
    \item Swahili
    \item Arabic
    \item Chinese
    \item English
    \item Ukrainian
\end{enumerate}

The idea here was simple: if the H-model is truly capable of adjusting its computation pathways depending on the input data, then it should learn to use different routing strategies for different languages—effectively generating distinct architectures on-the-fly for each one. This experiment also serves as a proxy for evaluating whether the model can learn to specialize portions of its architecture for different modalities or domains, even though in this case, we only worked with text.

The results were promising. In some training runs, the model displayed such distinct routing behaviors across languages that it was possible to infer the language of the input with ease just by looking at the routing visualization. In essence, the routing function started to act like a lightweight language detector, activating unique subpaths depending on the input. Unfortunately, some of the most striking examples of this effect were not saved during earlier runs. However, we still present examples of this behavior from distinct trained models that clearly exhibit this input-dependent architecture adaptation. The best example that we had on hand activated a distinct layer for each language, but this exact result is hard to redo. We trained several models that share this dynamic routing mechanism though. And these adaptive architectures are easily reproducible with such a task, but this always results in distinct routing patterns.

These results support our initial hypothesis: that the H-model, due to its modular and dynamic nature, is well-suited for tasks involving multiple data distributions or domains. It learns to specialize and reuse components across contexts, leveraging the shared hidden space rather than enforcing hard separation between processing pipelines. This behavior is particularly relevant in modern machine learning where multitask and multimodal capabilities are increasingly valuable.

This task, in particular, produced some of the most inspiring results in our study \ref{tab:multilingual_results}. Traditional models—both shallow and deep—struggled significantly on this multilingual language modeling task. A 1-layer Transformer and a 6-layer Transformer were each unable to learn meaningful representations, converging to only around 2\% accuracy with high loss values. In contrast, our H-model, even with modest architectural configurations, was able to achieve strong performance. Two H-model variants—one with 6 layers and 1 iteration, and another with 4 layers and 4 iterations achieved nearly 75\% and 95\% accuracy, respectively. These results not only highlight the effectiveness of dynamic routing in processing diverse inputs but also suggest that the H-model is capable of capturing language-specific computation paths that static models cannot. 

We believe that this significant performance gap is due to the limited width of the model in the size of its embeddings and feed-forward layers, as was mentioned earlier. The traditional and our models are usually same in size if they have the same number of layers though.

That said, we must acknowledge that to fully validate these findings and generalize them to broader settings, a more comprehensive study with larger training corpora, longer training schedules, and a wider set of languages would be necessary. Nevertheless, even in this constrained setup, the performance gap is stark and worthy of further investigation.

\begin{table}
\centering
\caption{Multilingual language modeling results (6 languages).}
\begin{tabular}{lcccccc}
\toprule
\textbf{Model} & \textbf{Layers} & \textbf{Iter} & $\alpha$ & $\alpha_{rate}$ & \textbf{ACC} & \textbf{Loss} \\
\midrule
${Tr}_2$ & 1 & - & - & - & 0.016 & 10.22 \\
${Tr}_2$ & 6 & - & - & - & 0.020 & 10.22 \\
${Hm}_2$ & 8(6) & 1 & 1 & 1 & \textbf{0.743} & \textbf{0.97} \\
${Hm}_2$ & 6(4) & 4 & 1 & 1 & \textbf{0.956} & \textbf{0.40} \\
\bottomrule
\end{tabular}
\label{tab:multilingual_results}
\end{table}

\begin{figure}
    \centering
    \begin{subfigure}[t]{0.45\textwidth}
        \includegraphics[width=\linewidth]{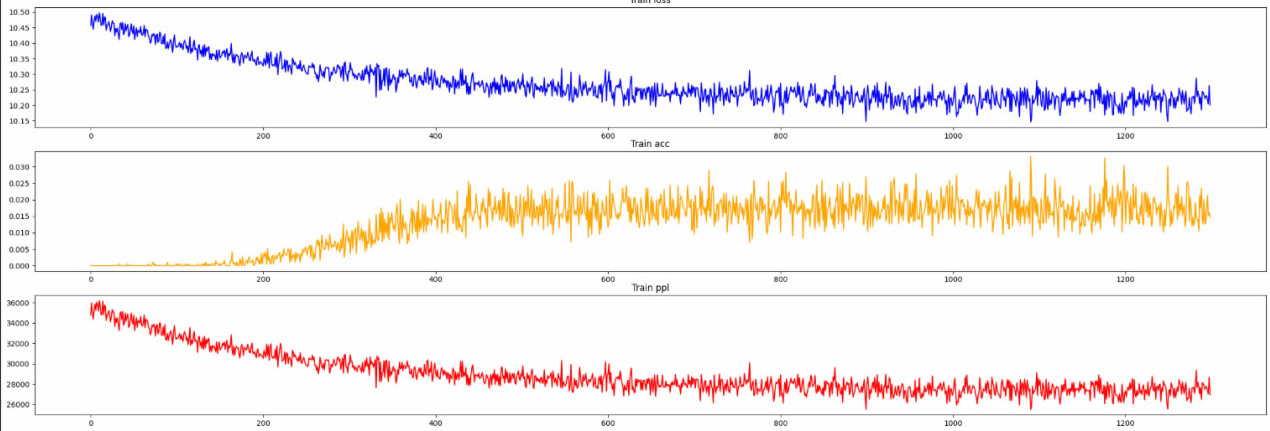}
        \caption{1 layer traditional model.}
    \end{subfigure}
    \hfill
    \begin{subfigure}[t]{0.45\textwidth}
        \includegraphics[width=\linewidth]{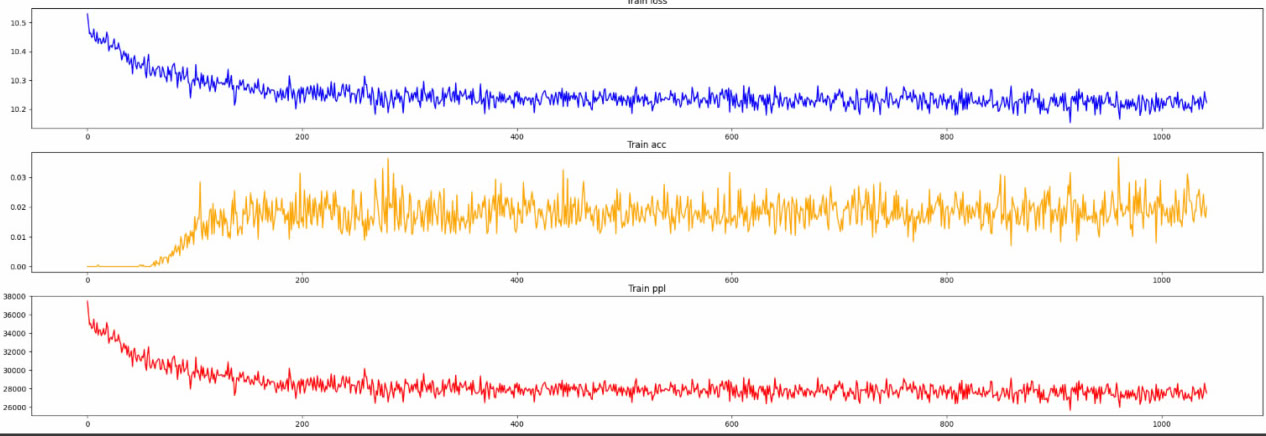}
        \caption{6 layers traditional model.}
    \end{subfigure}
    \caption{Plots of training of traditional models with 1 and 6 layers on multilingual dataset.}
    \label{fig:traditional_models_training_plots}
\end{figure}

To prove that traditional models actually converged, bellow are the plots of their training
\ref{fig:traditional_models_training_plots}.

\begin{figure}
    \centering
    \begin{subfigure}[t]{0.45\textwidth}
        \includegraphics[width=\linewidth]{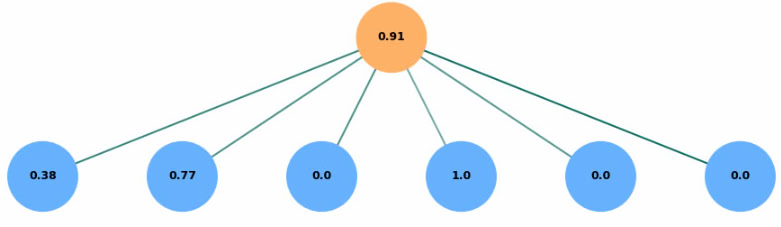}
        \caption{Ukrainian}
    \end{subfigure}
    \vspace{20pt}
    \hfill
    \begin{subfigure}[t]{0.45\textwidth}
        \includegraphics[width=\linewidth]{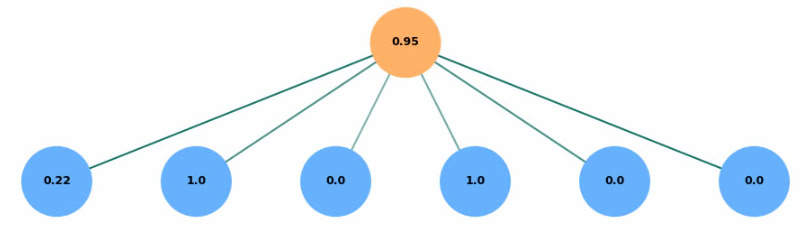}
        \caption{Chinese}
    \end{subfigure}
    \vspace{20pt}
    \hfill
    \begin{subfigure}[t]{0.45\textwidth}
        \includegraphics[width=\linewidth]{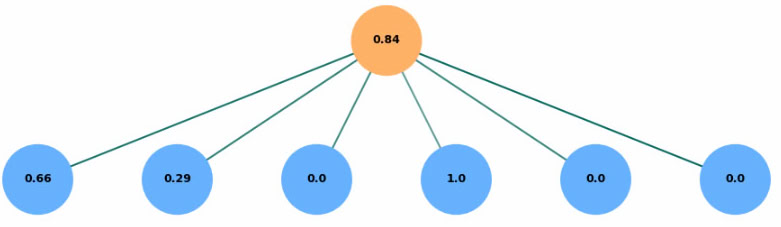}
        \caption{Swahili}
    \end{subfigure}
    \vspace{20pt}
    \hfill
    \begin{subfigure}[t]{0.45\textwidth}
        \includegraphics[width=\linewidth]{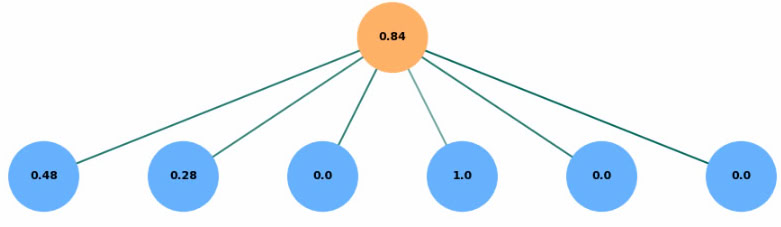}
        \caption{German}
    \end{subfigure}
    \vspace{20pt}
    \hfill
    \begin{subfigure}[t]{0.45\textwidth}
        \includegraphics[width=\linewidth]{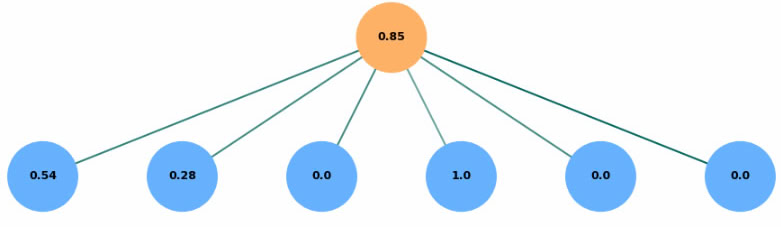}
        \caption{English}
    \end{subfigure}
    \caption{Architectural pathes of H-model with 6 H layers and 1 T iteration trained on multilingual language modeling task.}
    \label{fig:multilingual_h_model_6_1_visualizations}
\end{figure}

To continue the study we also share the examples of architectural routings of the H-models \ref{fig:multilingual_h_model_6_1_visualizations}. First we wanted to compare architectural patterns of the 1 T model. As we noticed, usually, layers are consistently more activated for 2 languages, Ukrainian and Chinese. This behaviour is highly reproducible and we believe that this is due to language differences and in particular total difference in alphabets and thus in embeddings, and that also makes those languages easily separable in the forward pass.

\begin{figure}
    \centering
    \begin{subfigure}[t]{0.45\textwidth}
        \includegraphics[width=\linewidth]{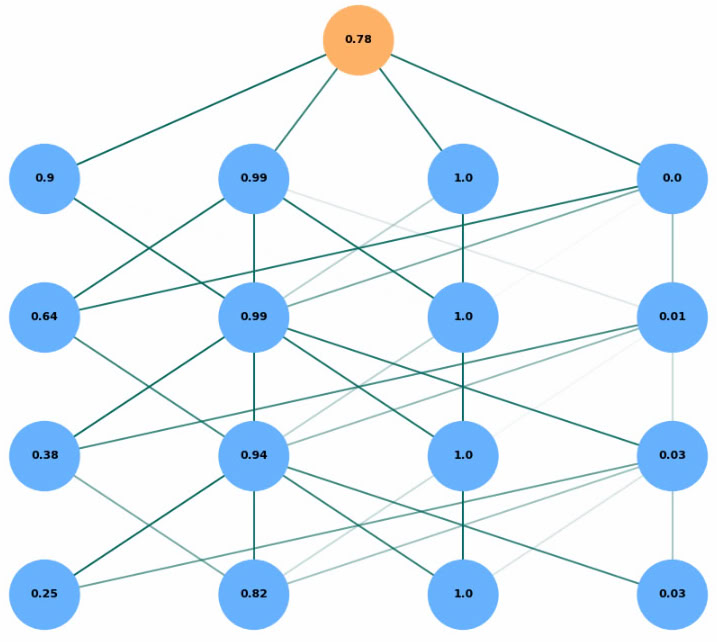}
    \end{subfigure}
    \hfill
    \begin{subfigure}[t]{0.45\textwidth}
        \includegraphics[width=\linewidth]{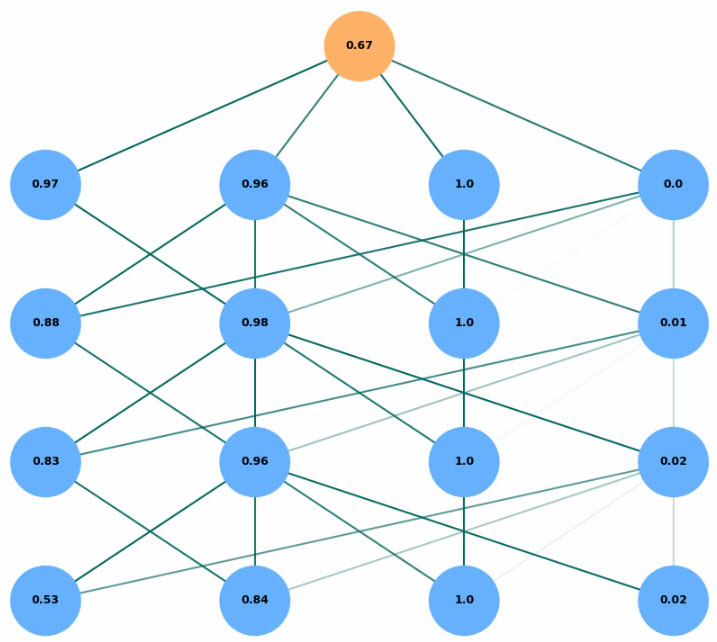}
    \end{subfigure}
    \hfill
    \begin{subfigure}[t]{0.45\textwidth}
        \includegraphics[width=\linewidth]{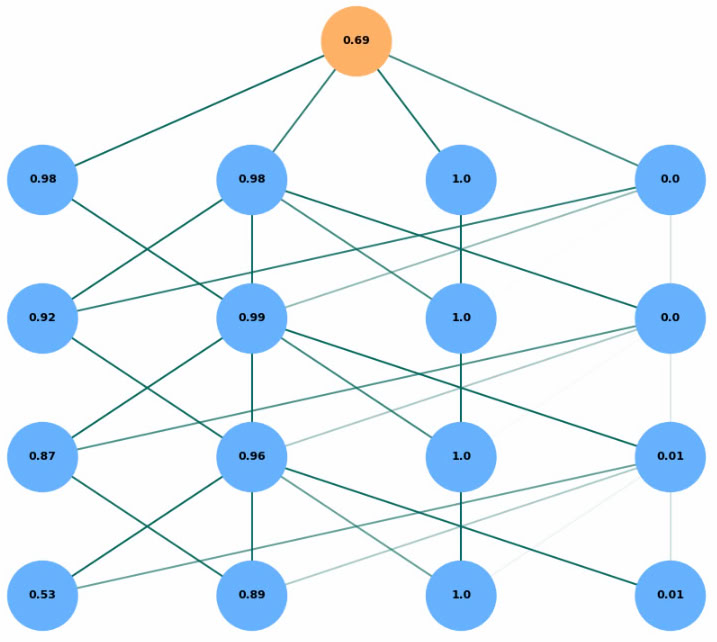}
    \end{subfigure}
    \hfill
    \caption{\textbf{German}}
    \label{fig:multilingual_h_model_4_4_visualizations_ge}
\end{figure}

\begin{figure}
    \centering
    \begin{subfigure}[t]{0.45\textwidth}
        \includegraphics[width=\linewidth]{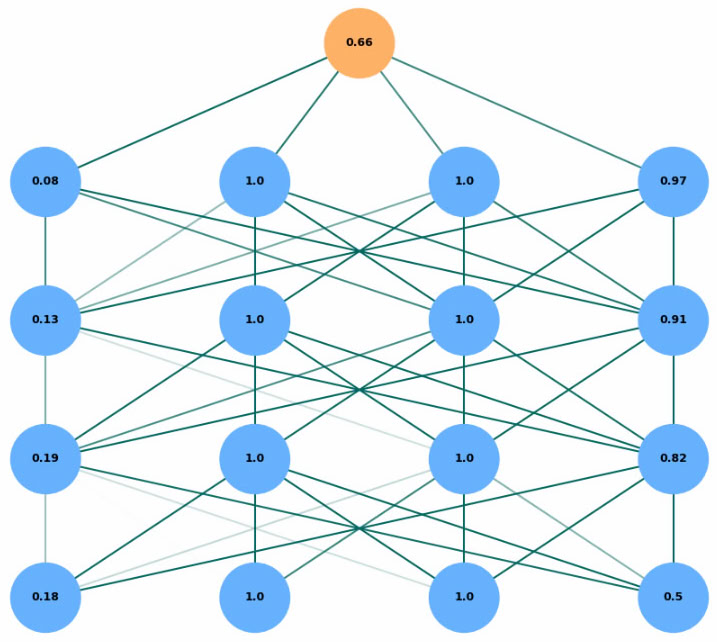}
    \end{subfigure}
    \hfill
    \begin{subfigure}[t]{0.45\textwidth}
        \includegraphics[width=\linewidth]{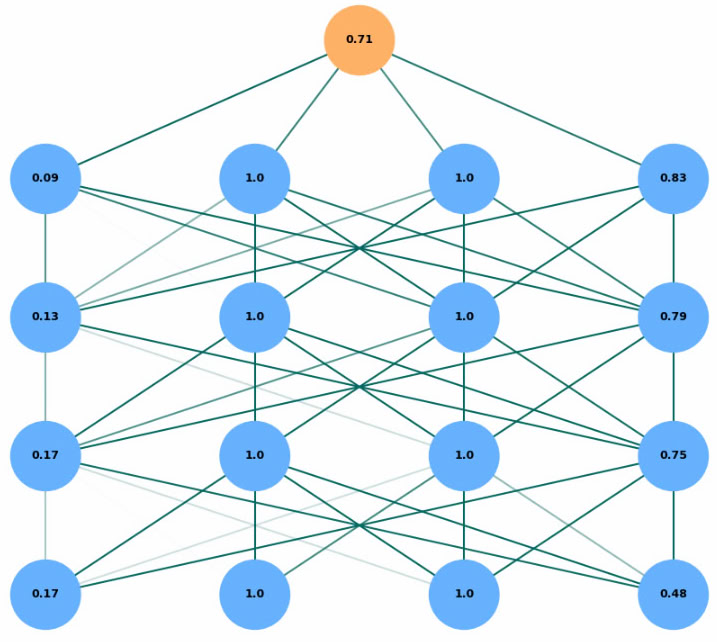}
    \end{subfigure}
    \hfill
    \begin{subfigure}[t]{0.45\textwidth}
        \includegraphics[width=\linewidth]{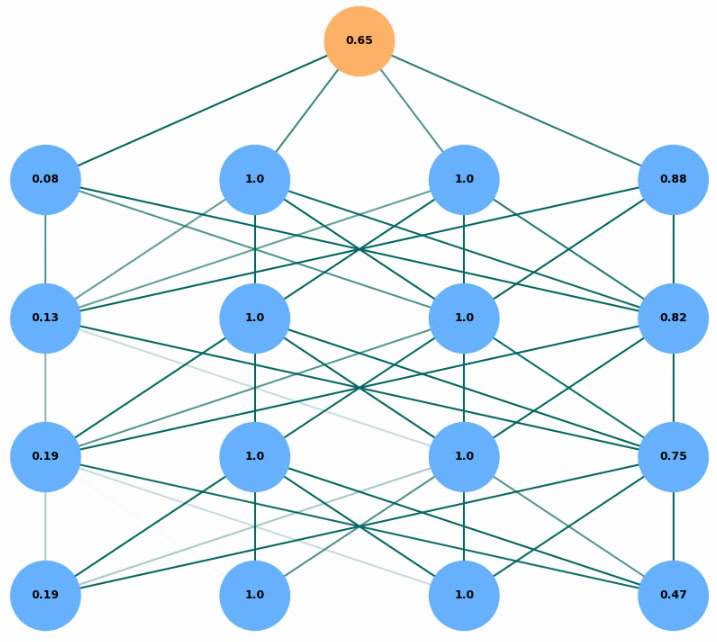}
    \end{subfigure}
    \hfill
    \caption{\textbf{Chinese}}
    \label{fig:multilingual_h_model_4_4_visualizations_ch}
\end{figure}

\begin{figure}
    \centering
    \begin{subfigure}[t]{0.45\textwidth}
        \includegraphics[width=\linewidth]{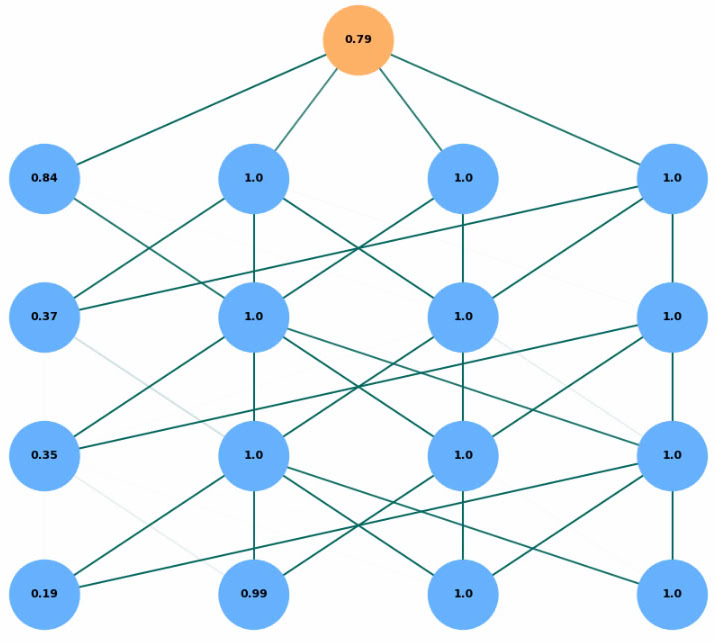}
    \end{subfigure}
    \hfill
    \begin{subfigure}[t]{0.45\textwidth}
        \includegraphics[width=\linewidth]{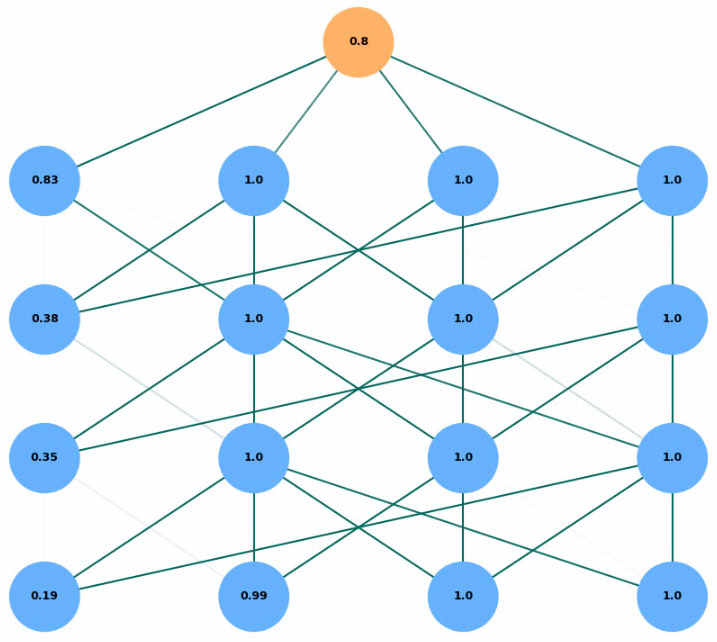}
    \end{subfigure}
    \hfill
    \begin{subfigure}[t]{0.45\textwidth}
        \includegraphics[width=\linewidth]{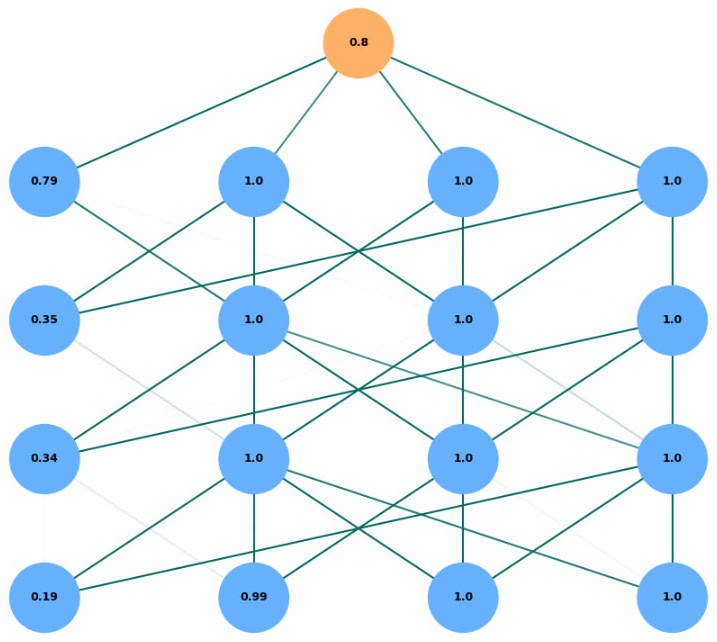}
    \end{subfigure}
    \hfill
    \caption{\textbf{Ukrainian}}
    \label{fig:multilingual_h_model_4_4_visualizations_uk}
\end{figure}

\begin{figure}
    \centering
    \begin{subfigure}[t]{0.45\textwidth}
        \includegraphics[width=\linewidth]{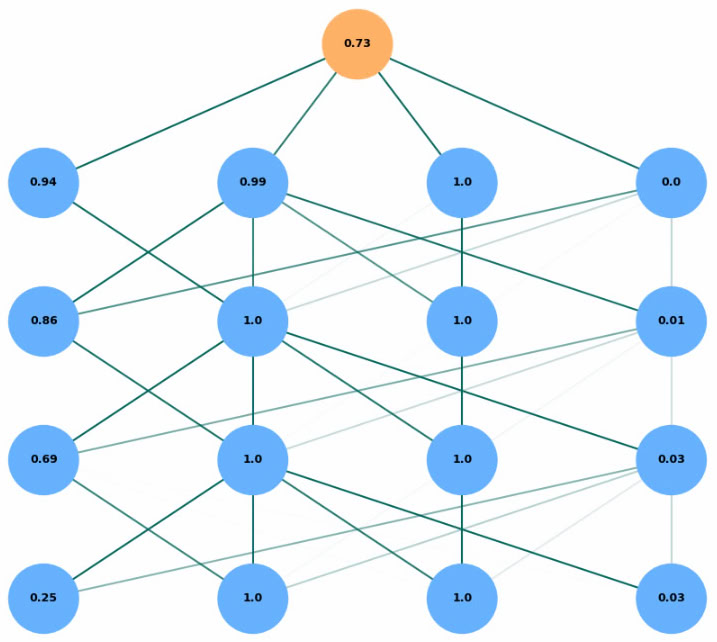}
    \end{subfigure}
    \hfill
    \begin{subfigure}[t]{0.45\textwidth}
        \includegraphics[width=\linewidth]{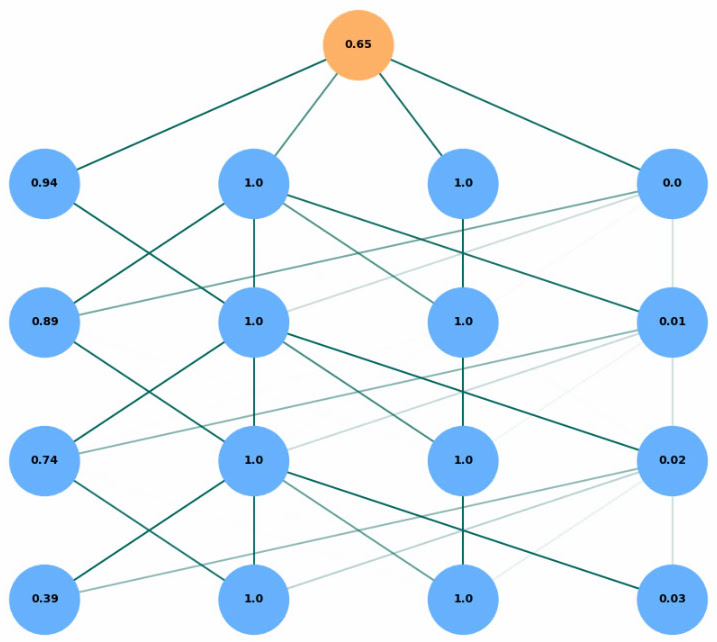}
    \end{subfigure}
    \hfill
    \begin{subfigure}[t]{0.45\textwidth}
        \includegraphics[width=\linewidth]{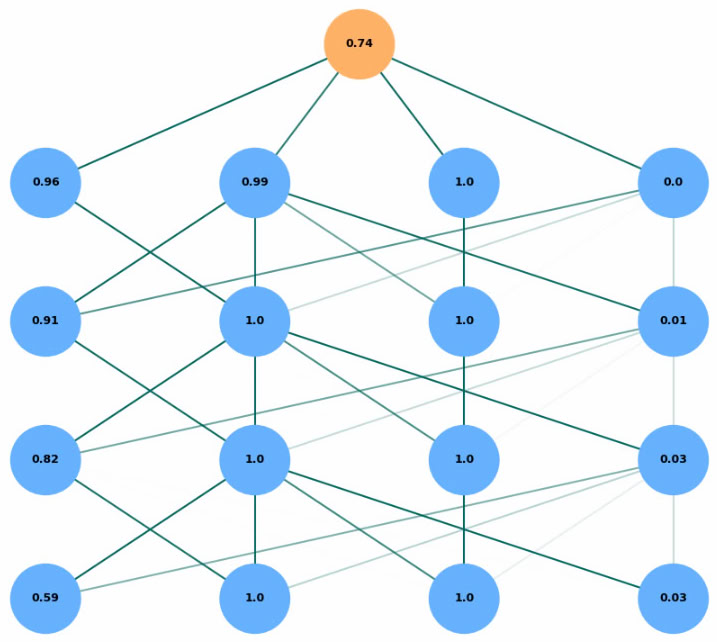}
    \end{subfigure}
    \hfill
    \caption{\textbf{English}}
    \label{fig:multilingual_h_model_4_4_visualizations_en}
\end{figure}

\begin{figure}
    \centering
    \begin{subfigure}[t]{0.45\textwidth}
        \includegraphics[width=\linewidth]{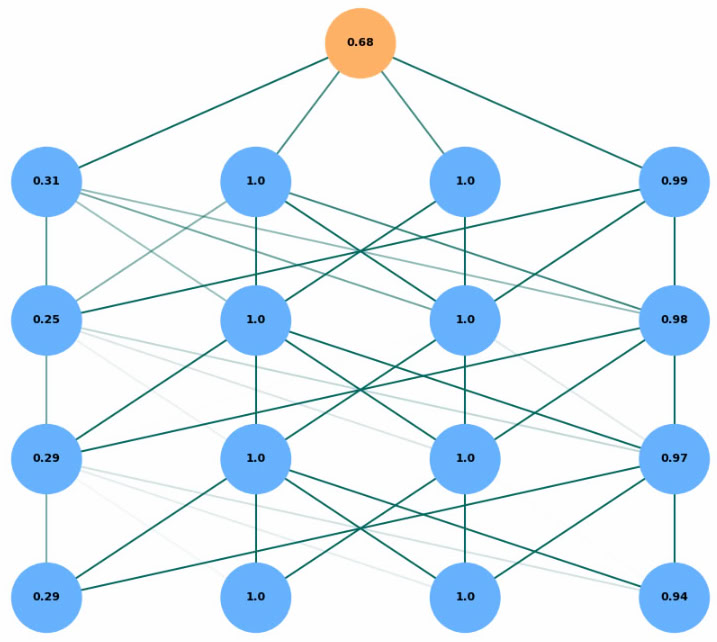}
    \end{subfigure}
    \hfill
    \begin{subfigure}[t]{0.45\textwidth}
        \includegraphics[width=\linewidth]{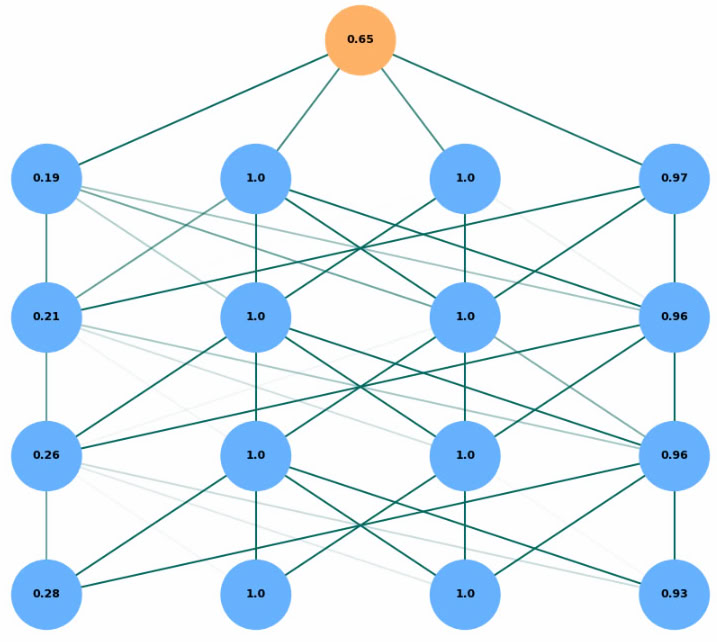}
    \end{subfigure}
    \hfill
    \begin{subfigure}[t]{0.45\textwidth}
        \includegraphics[width=\linewidth]{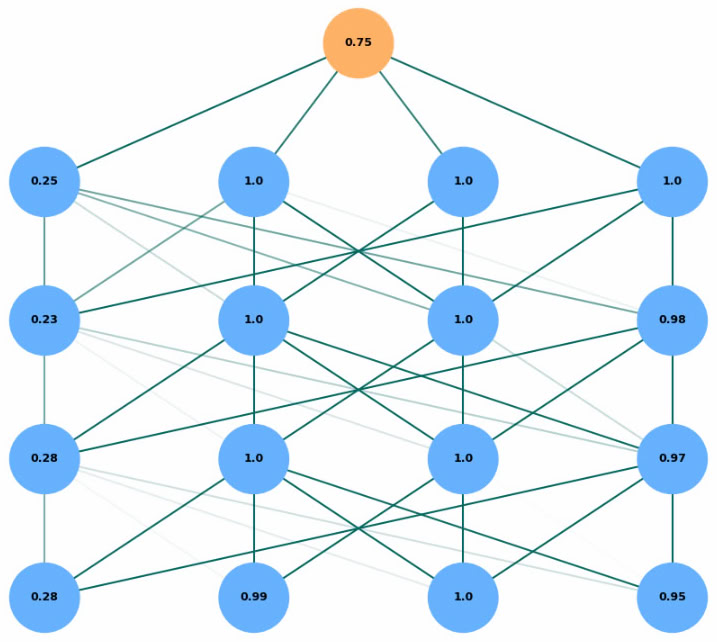}
    \end{subfigure}
    \hfill
    \caption{\textbf{Arabic}}
    \label{fig:multilingual_h_model_4_4_visualizations_ar}
\end{figure}

\begin{figure}
    \centering
    \begin{subfigure}[t]{0.45\textwidth}
        \includegraphics[width=\linewidth]{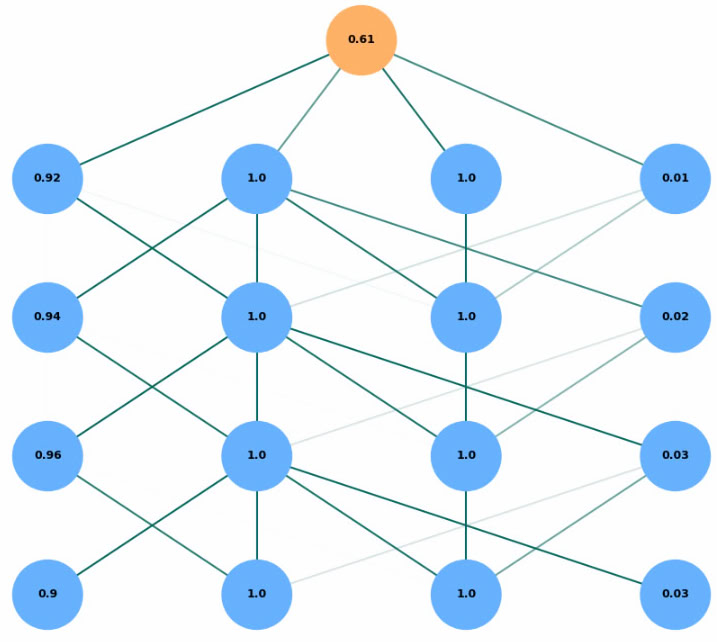}
    \end{subfigure}
    \hfill
    \begin{subfigure}[t]{0.45\textwidth}
        \includegraphics[width=\linewidth]{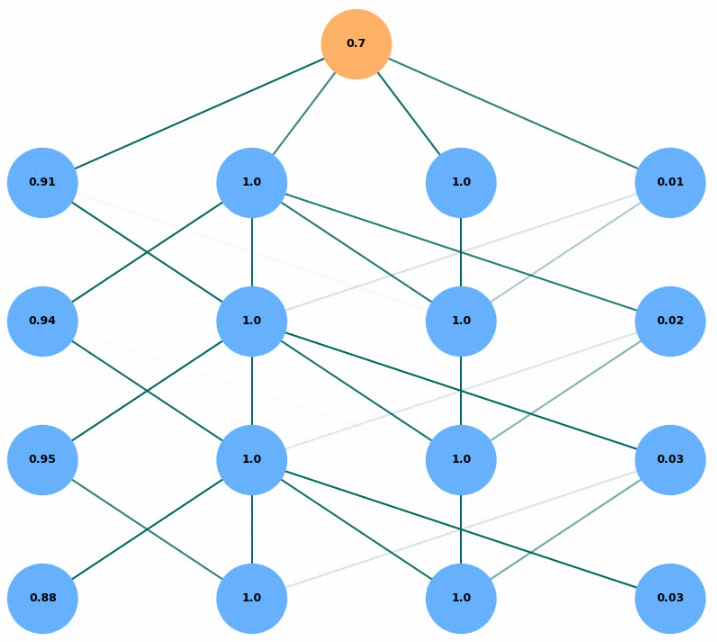}
    \end{subfigure}
    \hfill
    \begin{subfigure}[t]{0.45\textwidth}
        \includegraphics[width=\linewidth]{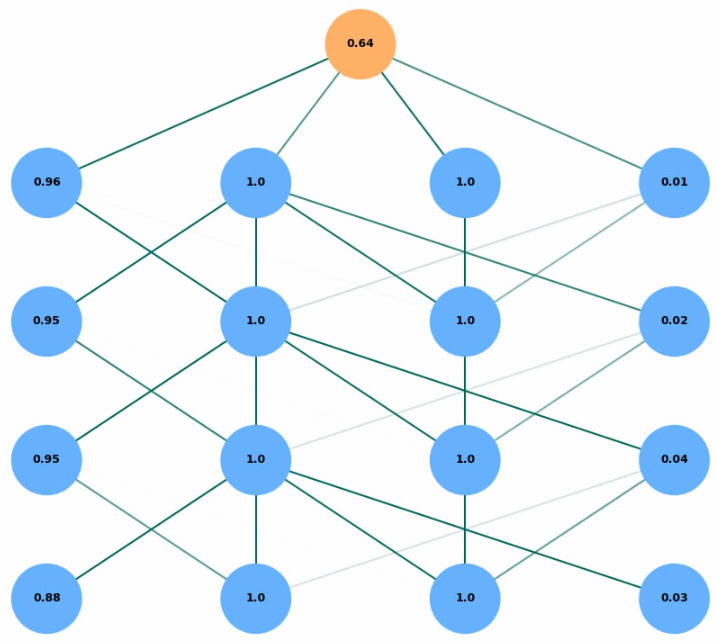}
    \end{subfigure}
    \hfill
    \caption{\textbf{Swahili}}
    \label{fig:multilingual_h_model_4_4_visualizations_sw}
\end{figure}

With deeper iterations the model shows much more complex routing patterns. Still we want to emphasize that those routings are stable across the same type of input, in this case, language. That is easily visible in the following visualizations of the H-model with 4 H layers and 4 T iterations architecture [ \ref{fig:multilingual_h_model_4_4_visualizations_ge}, \ref{fig:multilingual_h_model_4_4_visualizations_ch}, \ref{fig:multilingual_h_model_4_4_visualizations_uk}, \ref{fig:multilingual_h_model_4_4_visualizations_en}, \ref{fig:multilingual_h_model_4_4_visualizations_ar}, \ref{fig:multilingual_h_model_4_4_visualizations_sw} ].

Those results show that the routing pattern differs for input languages while staying consistent within the same language.

\subsection{Leveraging Combined Pretrained Models}

In this experiment, we explored the integration of multiple pretrained models within the H-model framework to assess their combined effectiveness on the Ohsumed classification task. Specifically, we incorporated seven layers from RoBERTa \cite{roberta} and three layers from MiniLM \cite{minilm} into the hidden layers of the H-model. The selected layers from RoBERTa were [0, 1, 2, 3, 5, 8, 12], and from MiniLM \cite{minilm} were [0, 2, 6]. To ensure unbiased routing decisions, the input and output layers were newly initialized, preventing any predisposition towards a particular pretrained model, given that layers within a single pretrained model are already optimized to communicate effectively.

To harmonize the differing embedding dimensions of RoBERTa \cite{roberta} and MiniLM \cite{minilm}, we employed a wrapper mechanism that projects their outputs into a unified embedding space, facilitating seamless integration within the H-model architecture.

Training was conducted with 10 T iterations, and the performance was benchmarked against a newly initialized traditional Transformer model and existing results on the Ohsumed dataset. While our model did not achieve state-of-the-art performance on the test set, it significantly outperformed the newly initialized model, indicating the potential benefits of integrating multiple pretrained models. Notably, the H-model exhibited signs of overfitting, achieving high training accuracy but comparatively lower test accuracy. This suggests that with appropriate regularization techniques and access to more extensive training data, the model's generalization capabilities could be enhanced.

An intriguing observation was the apparent communication between the two pretrained models within the H-model. However, as the number of layers and iterations increased, the complexity of the routing visualizations escalated, making it challenging to derive clear interpretations. 

One interesting thing we noticed across multiple experiments—including this one—is that layers often choose to send their outputs back to themselves. While this kind of behavior makes sense in fully trainable, randomly initialized networks, we didn’t really expect it to happen here, since the pretrained models we used already learned to communicate across their own layers in a predefined sequential way. This might suggest that the reason for that is deeper: it seems like layers are trying to stay relevant by keeping themselves active, maybe because the more they're used, the more gradient signal they get. In a way, it's like a kind of survival instinct within the model—layers that stay useful keep getting stronger, while others fade out. It’s a subtle but important dynamic that we see pop up in many of our setups, especially when routing is involved.

\begin{table}[h!]
\caption{Performance on the Ohsumed Classification Task. Tr with 12 layers actually failed to converge at all, thus the acc to this model is so low we decided not to include it in a table. This also means that we don’t suggest focusing solely on the result table, as the experiments were limited and far from optimal—instead, the main value lies in the observations and insights that emerged during the process.}
\begin{tabular}{lcccc}
\toprule
\textbf{Model} & \textbf{Layers} & \textbf{Iter} & \textbf{Train ACC} & \textbf{Test ACC} \\
\midrule
Tr & 1 & - & 0.55 & 0.12 \\
Tr & 12 & - & - & - \\
Hm & 10 (12) & 10 & 0.87 & 0.27 \\
\bottomrule
\end{tabular}
\label{tab:ohsumed_results}
\end{table}

These preliminary findings underscore the potential of the H-model to integrate and leverage multiple pretrained models effectively. However, to fully realize and validate this potential, further studies with enhanced training regimes and more substantial datasets are necessary \ref{tab:ohsumed_results}.

\subsection{Different Input Modalities and Mixed Architectures}

To further investigate the flexibility of the H-model in handling diverse input modalities, we extended our experiments to include both image and text data. Two tasks were selected to explore this setup. The first was the MM-IMDb dataset, where the model receives both an image (movie poster) and a plot summary and is asked to predict the genre. The second task was inspired by one of the CLIP \cite{clip} training objective: the model receives an image and a caption and must predict whether they match. We also tried a variant where the model is given two text captions instead of an image-caption pair.

Though these experiments were not fully optimized due to time and compute limitations, they exposed several meaningful behaviors in the H-model. Most notably, they revealed challenges tied to the aggregation of hidden states—previously identified as a potential bottleneck. This issue seems particularly relevant in the multimodal case, where modality-specific features need to be merged effectively across different network branches. Although we believe that this issue is still solved by our architecture, in a way that the layers can by themselves decide the routing pathes and thus create separate pathways for different modalities untill the model is ready to combine them. Creating separate pathways for different input modalities is a case that is highly applicable in modern architectures, but our model takes it further because it is only a subset of possible forward pathes variants of our architecture.

\begin{figure}
    \centering
    \begin{subfigure}[t]{0.45\textwidth}
        \includegraphics[width=\linewidth]{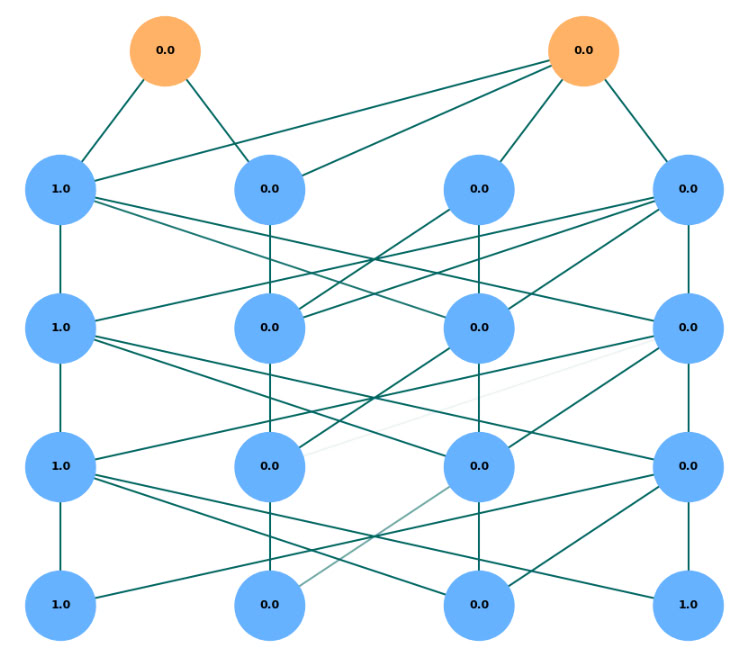}
        \caption{$alpha$=0.01}
    \end{subfigure}
    \hfill
    \begin{subfigure}[t]{0.45\textwidth}
        \includegraphics[width=\linewidth]{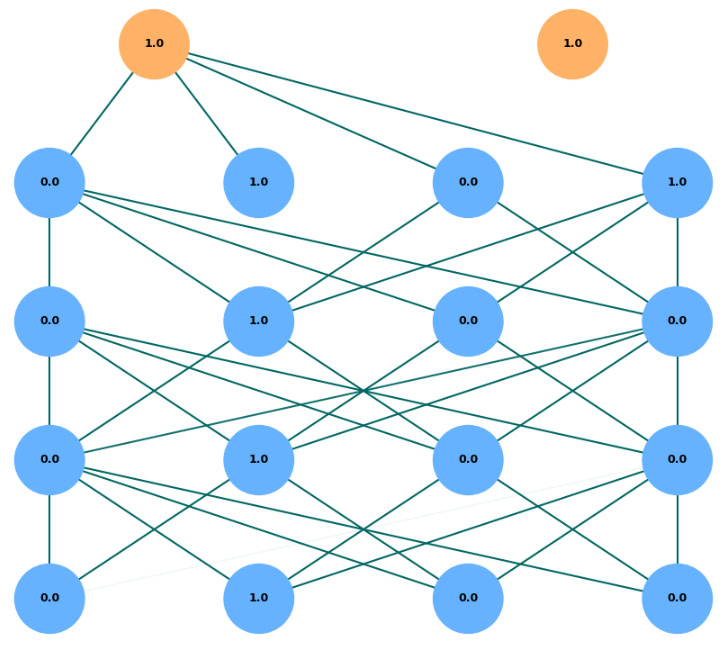}
        \caption{$alpha$=0.1}
    \end{subfigure}
    \hfill
    \begin{subfigure}[t]{0.45\textwidth}
        \includegraphics[width=\linewidth]{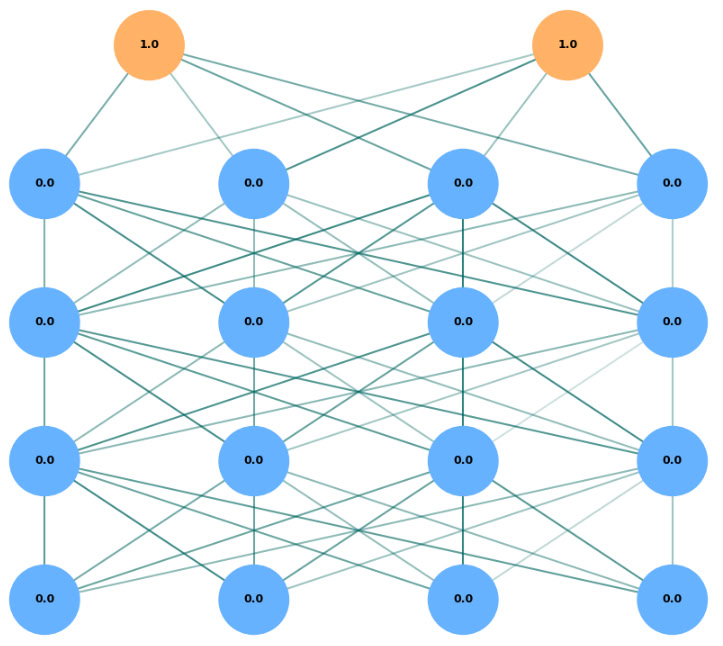}
        \caption{$alpha$=1}
    \end{subfigure}
    \hfill
    \caption{Architectures of H-models trained on MM-IMDB with different $\alpha$ parameter set.}
    \label{fig:imdb_ht_alpha}
\end{figure}

One promising observation, however, was the role of the $\alpha$ hyperparameter. In scenarios where the value of $\alpha$ was reduced, the model tended to explore deeper iterations more extensively. This allowed the architecture to form more stable and meaningful routing paths, especially in cases where the input signal required longer computation chains. Some examples of how $\alpha$ influenced the model architecture are presented \ref{fig:imdb_ht_alpha}. 

As shown on Figures \ref{fig:imdb_ht_alpha} with $\alpha$=0.1 the model did not have connections from the second input layer that is responsible for images. This behaviour might be because of task specification, as it is possible to predict the label only based on the plot or the poster of the movie. That is the main reason why we took it further and trained several models for another task that was mentioned previously - image-caption matching, or caption-caption matching. With such a task the model would be able to predict the correct label only by taking into account both information flows.

With the second type of experiments we actually managed to prove that the model keeps the ability to processing several input modalities at once effectively. The caption-caption matching model actually introduced us to dynamic architectural routing one more time.

A lot more studies could be made in this direction though. Due to the interpretability of our model we can almost fully analyze the decisions of our models. This opens a huge room for experiments and opens a little of what is going on under the hood, not keeping everything in a black box as is known.

\subsection{Varying Architectures used for layers as a minimal block of H-model} 
\begin{figure}[h!]
    \centering
    \begin{subfigure}[t]{0.45\textwidth}
        \includegraphics[width=\linewidth]{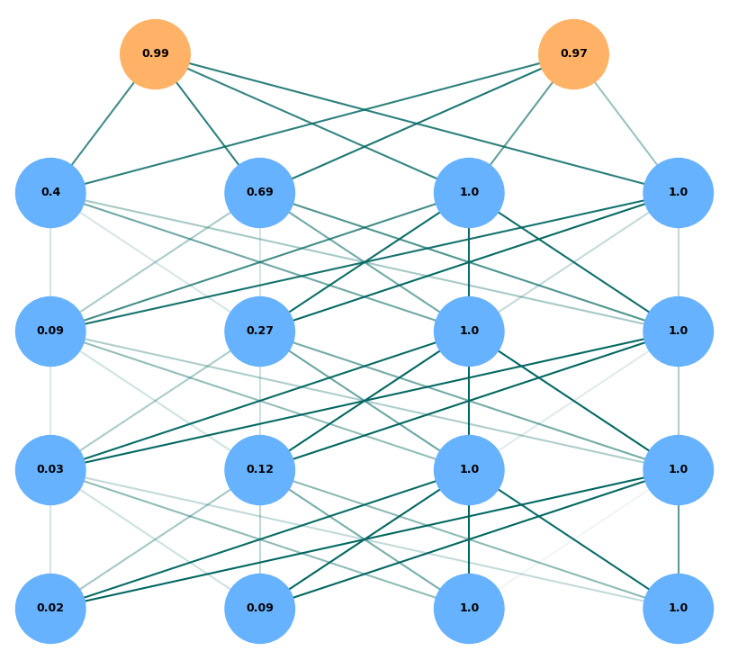}
    \end{subfigure}
    \hfill
    \caption{Architectural bias of the model due to layer specific architectures. In this example we used a lot simpler computation pattern for connections in convolutional layers than in transformer layers. The H-model architecture can be easily described as follows (the left part is transformers and the right part are convolutions with text and image inputs respectively). As you can see the convolutional layers have a lot simpler and constant patterns of routing.}
    \label{fig:imdb_h_trans_convs}
\end{figure}

In this setting, we also experimented with using different architectural types for the routing layers. Specifically, we included both convolutional layers and Transformer blocks as routing candidates. This introduced new complexity into the routing patterns, as the model had access to computationally distinct layer types. Interestingly, we observed that the behavior of each layer was highly influenced by the type of computations it was capable of performing. Transformer blocks typically formed more complex routing patterns, while convolutional layers (which were kept relatively simple due to implementation constraints) showed correspondingly simpler routing behaviors. This difference is clearly visible in the routing visualizations \ref{fig:imdb_h_trans_convs}.

Overall, this line of experimentation demonstrates the potential of the H-model to mix and match computational "flavors," where each layer can contribute a unique transformation path. As a result, the model has the capacity to adaptively route information through a variety of functional building blocks, making it especially interesting for future multimodal tasks.

\subsection{Results and Insights}

While the primary goal of this study was not to produce state-of-the-art performance on benchmark tasks, the experiments have nevertheless revealed a wide range of compelling and sometimes unexpected behaviors in the H-model. Below, we summarize the key insights that emerged across our evaluations:

\begin{itemize}
    \item \textbf{Stable Architectures Emerge Naturally:} In many tasks, the H-model converged to consistent routing patterns across all inputs. These stable architectures resemble traditional forward passes and demonstrate the model’s ability to self-organize into efficient computation paths.

    \item \textbf{Routing as Implicit Architecture Search:} The ability to learn routing paths dynamically allowed the H-model to function similarly to neural architecture search methods—automatically discovering computation strategies tailored to the task.

    \item \textbf{Adaptive architectures} H-model can dynamically change its architectural patterns based on the input. This is undeniably the strongest side of our model. We did not expect this property to appear so often, as experiments have shown. Almost all our models developed this property, leaning towards more diverse or stable routing pathes, depending on the case.

    \item \textbf{Gradient Flow Encourages Self-Routing:} Across multiple tasks, a common pattern was that layers often routed outputs back to themselves. We believe this stems from gradient dynamics—layers that route to themselves receive more gradient signal, leading to a form of layer-level competition or “survival mechanism.” Despite this surprising property, models do not limit their computations to one layer reuse, but are able to develop complex patterns.

    \item \textbf{Multilingual Task Shows Clear Adaptivity:} When trained on a six-language modeling task, the model developed language-specific routing paths. Some were so distinct that language identity could be visually inferred from the routing graph, suggesting emergent specialization without direct supervision.

    \item \textbf{Traditional Models Struggled Where H-Model Adapted:} On the multilingual task, traditional Transformer models failed to converge, while the H-model achieved over 95\% accuracy in some configurations, highlighting the potential of adaptive routing in diverse settings. We don't want to focus on that, though, because we believe that the experiments we provided cannot be fully reliable and need further investigation.

    \item \textbf{Combining Pretrained Models:} In our hybrid model trained on the Ohsumed classification task, we successfully integrated two different pretrained backbones (RoBERTa and MiniLM). The models managed to create meaningful interconnections between them. Though the pretrained hybrid model overfitted on limited data, this overfitting suggests an underlying capacity that could be harnessed with better training setups and regularization.

    \item \textbf{Routing Complexity Reflects Architecture Type:} When combining convolutional and Transformer layers, routing graphs showed architecture-dependent complexity. Transformers encouraged more complex routing patterns, while simpler conv layers led to sparse, regular paths. This is due to the implementation of connection computation in convolutional layers.

    \item \textbf{Layer Behavior Depends on Computation Type:} Beyond routing, each layer’s function appears shaped by its internal architecture. This highlights a unique advantage of the H-model: it enables heterogeneous computation where the model not only routes information adaptively but also chooses among qualitatively different computation styles.

    \item \textbf{Better endurance to Deeper Iterations than with just Residual connections:} In deeper H-models, convergence was often more reliable than in traditional deep networks. This suggests that routing mechanisms can alleviate training instability in deep architectures—a problem traditionally addressed by residual connections, but it is still not fully solved.

    \item \textbf{H-Model Skip Connections Are More Flexible:} Unlike residual networks, where skip paths are additive and always active, the H-model can learn to bypass layers entirely using routing weights. This allows information to "skip" layers in a more selective and interpretable way.

    \item \textbf{Visualization Provides Interpretability:} Throughout experiments, routing visualizations offered valuable interpretability. They allowed us to pinpoint where the model preferred deeper computation, when layers were reused, and how task complexity shaped path selection.

    \item \textbf{Alpha Hyperparameter Offers Behavioral Control:} Lowering the $\alpha$ parameter led to deeper routing behavior. This gives us a tool to modulate the computational depth adaptively without hard-coding depth into the architecture.

    \item \textbf{Aggregation Strategy is a Bottleneck in Multimodal Settings:} While the model handles multiple modalities in parallel, aggregating hidden states into a final output remains challenging and likely limits our performance in multimodal matching tasks. We believe that this is the most effective implementation that would not lead to an even longer forward pass, though. 

    \item \textbf{Input Diversity Challenges Static Models:} When the model was exposed to diverse inputs—whether languages or modalities—it adapted its structure in ways that static architectures could not, especially when trained from scratch.

    \item \textbf{Experiments Reveal More Than Metrics:} Due to resource limitations, not all experiments reached convergence or full performance. However, each setup offered valuable insights into architectural behavior, routing dynamics, and future research directions.
\end{itemize}

\subsection{Advantages and Limitations}

The proposed H-model architecture introduces several practical and conceptual benefits over traditional deep learning architectures, especially in dynamic and multi-modal settings. However, it also comes with a notable trade-off in efficiency. Below, we summarize both its primary advantages and its main limitation.

\begin{itemize}
    \item \textbf{Advantages:}
    \begin{itemize}
        \item \textbf{Dynamic Routing:} The model learns task-specific and input-adaptive computational paths, functioning as a built-in architecture search without needing manual design.
        
        \item \textbf{Heterogeneous Computation:} Supports the combination of diverse layer types (e.g., convolutions and transformers), allowing for richer and more specialized representations within a single architecture.
        
        \item \textbf{Improved Deep Network Stability:} The routing mechanism provides a more stable alternative to traditional residual connections, especially in deeper models that often suffer from vanishing gradients or convergence issues.
        
        \item \textbf{Model Interpretability:} Routing visualizations offer insight into how the model makes decisions, how different layers are utilized, and how computational depth varies per input or task.
        
        \item \textbf{Emergent Specialization:} Observed in multilingual and multimodal setups, where the model developed distinct routing paths—effectively specializing substructures without explicit supervision. This differs from the dynamic architecture in the sense that some layers are activated only with a specific type of input, specializing in that type of input.
        
        \item \textbf{Cross-Model Integration:} Capable of combining multiple pretrained models into a coherent, communicating architecture, extending transfer learning beyond individual networks.
    \end{itemize}

    \item \textbf{Limitation:}
    \begin{itemize}
        \item \textbf{Slower Forward Propagation:} Due to iterative routing and dynamic path evaluation, the forward pass is significantly slower than in traditional static architectures. This affects training and inference speed, particularly on resource-constrained setups or tasks requiring high throughput. If we assume that we always want to use the same number of iterations as a number of layers, then the forward pass time is squared compared to traditional models, and this is critical in the modern competitive world with huge trillion-parameter models. But we suggest to use one of the other two proposed possible settings when we have more layers than iterations, or the other way around. We believe that the assumption is that the model need the same size of memory (number of layers) as intelligence (number of iterations). This is rather specific to the task at hand. But in comparison to modern models that are more specialized, we suggest the setting with fewer layers than iterations.
    \end{itemize}
\end{itemize}

\section{Conclusion}

In this work, we introduced and explored the H-model architecture—a dynamic, iterative, and flexible framework designed to adapt its computational path per input and task. Across a diverse set of experiments ranging from language modeling to multimodal reasoning and model fusion, we demonstrated that the H-model is capable of forming stable and interpretable computation patterns, leveraging both constant and adaptive architectures.

While our experiments were limited in scope due to computational constraints, they revealed a number of promising behaviors: emergent modularity, dynamic specialization, and improved depth stability. These findings suggest that the H-model has strong potential as a general-purpose framework for building interpretable, adaptable, and compositional neural networks. Also, as you might have noticed from the visualization of H-model architectures, the typical pattern of the traditional forward pass was never reproduced with this type of architecture, suggesting that the conventional methods are far from optimal.

Future work will aim to refine the routing mechanism, improve efficiency, and further test the model on more complex and large-scale multimodal tasks. Additionally, we plan to develop a new pretraining objective that is specifically designed for the H-model’s architecture. This objective would be inherently tied to the routing and compositional capabilities of the model and may offer performance advantages that are unattainable with conventional training paradigms.

\bibliography{references}

\end{document}